%% file: main.tex
\documentclass[10pt,journal,compsoc]{IEEEtran}
\ifCLASSOPTIONcompsoc
  \usepackage[nocompress]{cite}
\else
  \usepackage{cite}
\fi
\ifCLASSINFOpdf
\else
\fi
\usepackage[whole]{bxcjkjatype} 

\RequirePackage{plautopatch}
\usepackage{graphicx}
\usepackage{amsmath}
\usepackage{amssymb}
\usepackage{booktabs}
\usepackage{tikz}
\usetikzlibrary{positioning}
\usetikzlibrary{arrows.meta}
\usetikzlibrary{shapes.callouts}

\usepackage[capitalize]{cleveref}

\makeatletter
\let\MYmakecaption\@makecaption
\makeatother
\usepackage{subcaption}
\makeatletter
\let\@makecaption\MYmakecaption
\makeatother

\definecolor{sxfer-red}{RGB}{173,0,0}
\definecolor{sxfer-blue}{RGB}{0,0,173}

\begin{document}
%
\title{Style Feature Extraction Using Contrastive Conditioned Variational Autoencoders with Mutual Information Constraints}
%
%
%
%

\author{Suguru~Yasutomi,~\IEEEmembership{Member,~IEEE} and
        Toshihisa~Tanaka,~\IEEEmembership{Senior~Member,~IEEE}
\IEEEcompsocitemizethanks{\IEEEcompsocthanksitem S. Yasutomi and T. Tanaka with the Department of Electronic and Information Engineering, Tokyo University of Agriculture and Technology, 2--24--16 Nakacho, Koganei-shi, Tokyo,
  184--8588 Japan\protect\\
}
\thanks{This work was supported by CREST (JPMJCR1784) of the Japan Science and Technology Agency (JST).}
}

\IEEEtitleabstractindextext{%
\begin{abstract}
Extracting fine-grained features such as styles from unlabeled data is crucial for data analysis.
Unsupervised methods such as variational autoencoders (VAEs) can extract styles that are usually mixed with other features.
Conditional VAEs (CVAEs) can isolate styles using class labels; however, there are no established methods to extract only styles using unlabeled data.
In this paper, we propose a CVAE-based method that extracts style features using only unlabeled data.
The proposed model consists of a contrastive learning (CL) part that extracts style-independent features and a CVAE part that extracts style features.
The CL model learns representations independent of data augmentation, which can be viewed as a perturbation in styles, in a self-supervised manner.
Considering the style-independent features from the pretrained CL model as a condition, the CVAE learns to extract only styles.
Additionally, we introduce a constraint based on mutual information between the CL and VAE features to prevent the CVAE from ignoring the condition.
Experiments conducted using two simple datasets, MNIST and an original dataset based on Google Fonts, demonstrate that the proposed method can efficiently extract style features.
Further experiments using real-world natural image datasets were also conducted to illustrate the method's extendability.
\end{abstract}

}

\maketitle

\IEEEdisplaynontitleabstractindextext

%
\IEEEpeerreviewmaketitle

\input{introduction}

\input{relatedwork}

\input{method}

\input{experiments}

\input{discussion}

\input{conclusions}

%

%
%

\ifCLASSOPTIONcaptionsoff
  \newpage
\fi



\bibliographystyle{IEEEtran}
\bibliography{IEEEabrv,library}

%

\begin{IEEEbiography}[{\includegraphics[width=1in,height=1.25in,clip,keepaspectratio]{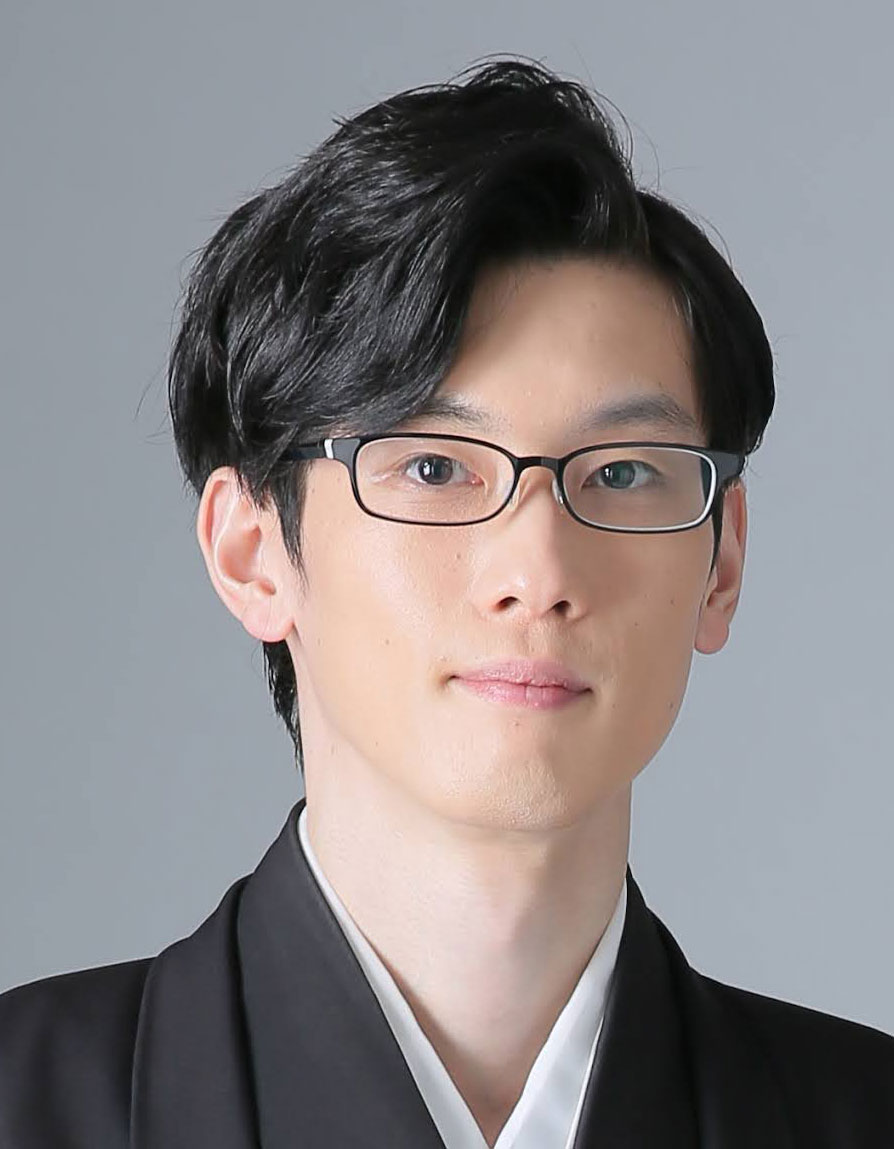}}]{Suguru Yasutomi}
received the B.E. and M.E. degrees from the Tokyo University of Agriculture and Technology, Tokyo, Japan in 2014 and 2016, respectively.
He is currently working on his Ph.D. in electronic and information engineering at the same university.
He joined Fujitsu in 2016 and has been working on the research of artificial intelligence including deep learning.
His research interests are machine learning and signal processing, especially with small or noisy data.
\end{IEEEbiography}

\begin{IEEEbiography}[{\includegraphics[width=1in,height=1.25in,clip,keepaspectratio]{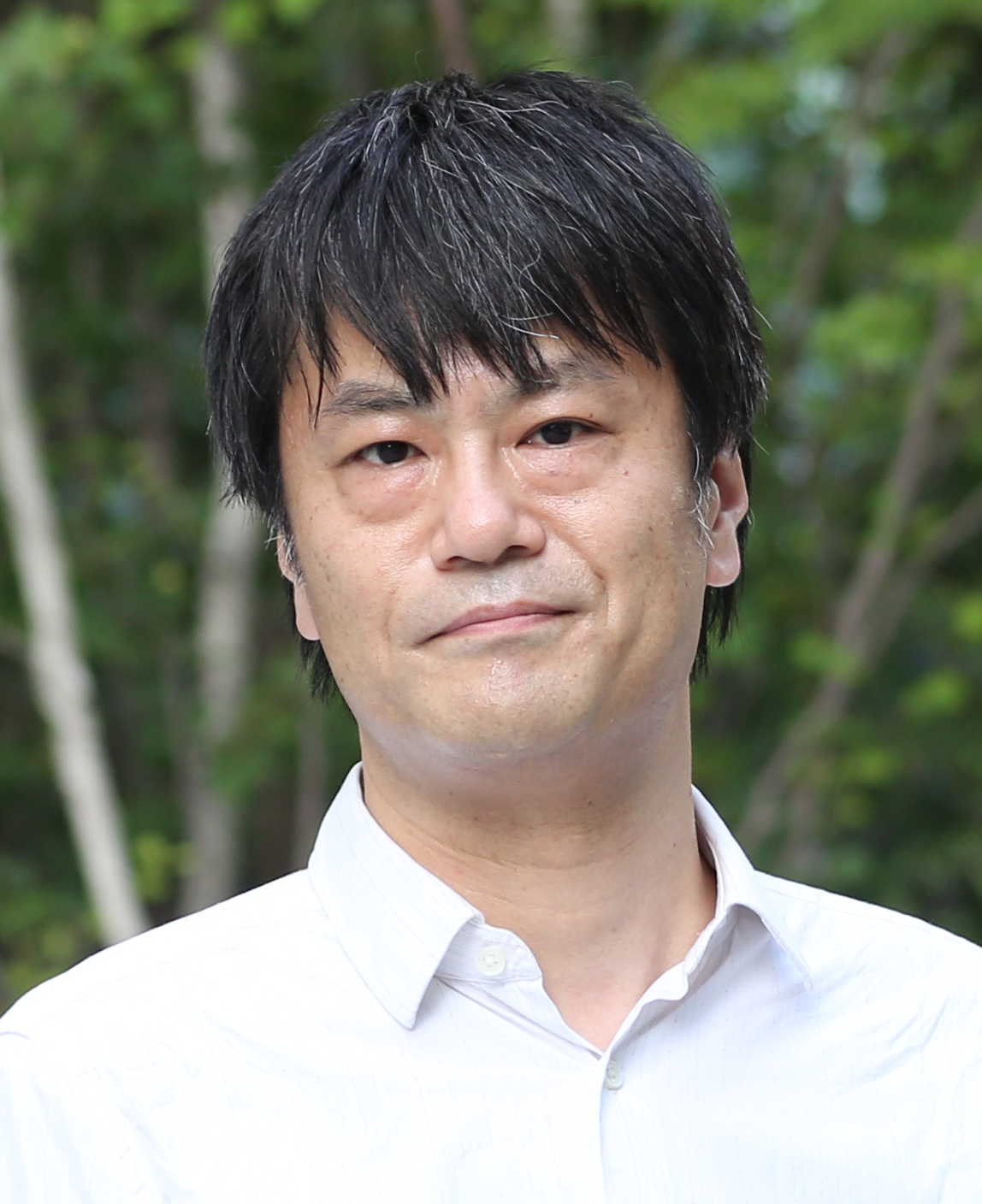}}]{Toshihisa Tanaka}
received the B.E., M.E., and Ph.D. degrees from the Tokyo Institute of Technology in 1997, 2000, and 2002, respectively. From 2000 to 2002, he was a JSPS Research Fellow. From October 2002 to March 2004, he was a Research Scientist at RIKEN Brain Science Institute. In April 2004, he joined the Department of Electrical and Electronic Engineering, at the Tokyo University of Agriculture and Technology, where he is currently a Professor. In 2005, he was a Royal Society visiting fellow at the Communications and Signal Processing Group, Imperial College London, U.K. From June 2011 to October 2011, he was a visiting faculty member in the Department of Electrical Engineering, the University of Hawaii at Manoa.

His research interests include a broad area of signal processing and machine learning, including brain and biomedical signal processing, brain-machine interfaces, and adaptive systems. He is a co-editor of Signal Processing Techniques for Knowledge Extraction and Information Fusion (with Mandic, Springer), 2008, and a leading co-editor of Signal Processing and Machine Learning for Brain-Machine Interfaces (with Arvaneh, IET, U.K.), 2018.

He served as an associate editor and a guest editor of special issues in journals, including IEEE Access, Neurocomputing, IEICE Transactions on Fundamentals, Computational Intelligence and Neuroscience (Hindawi), IEEE Transactions on Neural Networks and Learning Systems, Applied Sciences (MDPI), and Advances in Data Science and Adaptive Analysis (World Scientific). He served as editor-in-chief of Signals (MDPI). Currently, he serves as an associate editor of Neural Networks (Elsevier). Furthermore, he served as a Member-at-Large, on the Board of Governors (BoG) of the Asia-Pacific Signal and Information Processing Association (APSIPA). He was a Distinguished Lecturer of APSIPA. He serves as the Vice-President for Member Relations and Development of APSIPA. He is a senior member of IEEE, and a member of IEICE, APSIPA, the Society for Neuroscience, and the Japan Epilepsy Society. He is the co-founder and CTO of Sigron, Inc.
\end{IEEEbiography}







\end{document}

%% file: introduction.tex
\IEEEraisesectionheading{
\section{Introduction}\label{sec:introduction}
}

\IEEEPARstart{E}{xtracting} information or knowledge from data is a crucial step in engineering.
Tasks for extracting information from unlabeled data are called unsupervised learning~\cite{Bishop2006,Murphy2012}.
Traditional unsupervised learning methods, including dimensionality reduction methods, including principal component analysis (PCA), and clustering methods, such as $k$-means, have been extensively studied~\cite{Bishop2006,Murphy2012}.
In the recent decade, unsupervised learning using deep neural networks (DNNs) with large expression capability been used successfully for extracting information~\cite{Goodfellow2016}.

Autoencoders (AEs) are typical DNN-based unsupervised models~\cite{Vincent2008}.
AEs work like a nonlinear version of PCA and consist of encoders that map input data to feature spaces and decoders that reconstruct the input data from vectors in the feature spaces.
In DNNs, representation learning involves finding a map to beneficial feature spaces using unlabeled data.
Unsupervised learning is a typical example of representation learning.
Further, a self-supervised learning method called contrastive learning (CL) has recently gained popularity~\cite{Le-Khac2020}.

CL can learn representations suitable for classifying images~\cite{Le-Khac2020,SimCLR,MoCo,SimSiam}.
The underlying idea behind CL is that samples with similar contents should be mapped to the identical feature vector.
To achieve this, CL needs a large number of pairs characterized by similar contents; however, it is not easy to prepare such a dataset.
Thus, data augmentation is used for generating a pair of samples with the same content from each sample.
The CL model behaves as a feature extractor robust against perturbations introduced by data augmentation, such as blurring and changes in brightness and contrast.
In this way, even a simple linear model can efficiently classify samples in ImageNet~\cite{ILSVRC15} using CL for feature extraction.
Trained CL models are considered good initial models for object detection tasks~\cite{SimCLR,MoCo,SimSiam}.
How CL works with the simple data augmentation strategy remains an open question; however, some explanations have been reported.
For example, CL isolates contents from other miscellaneous features by learning features independent of data augmentation~\cite{Kugelgen2021}, and CL implicitly learns the inverse of the data generating process~\cite{Zimmermann2021}.

CL and AEs can extract features that are suitable for classification and dominant components of the given data, respectively.
On the other hand, it is also crucial to extract more fine-grained information, such as styles.
Styles are often referred to as features that do not correspond to the main contents of data (e.g., class labels), and they are important because they include unique characteristics such as individual differences, the environment where the data are obtained, and domains.
Well-known models for extracting style features include variational AEs (VAEs)~\cite{Kingma2014} and conditional VAEs (CVAEs)~\cite{Kingma2014a,Sohn2015}.
VAEs typically learn to construct feature spaces that represent rough features such as classes as clusters.
In each cluster, finer features like styles are also represented.
A CVAE is a variation of a VAE conditioned by labels, if available.
With CVAEs, feature spaces are learned to represent styles.
CVAEs trained with MNIST~\cite{MNIST}, which is a handwritten digit image dataset, forms a feature space with a distribution representing handwriting styles (e.g., thickness of lines and angle of digits)~\cite{Kingma2014a}.
Thus, CVAE is an efficient model for extracting only styles when labels are available.
However, little attention has been paid to a model without labels that can extract only styles.
In this paper, we establish a method for extracting style features from an unlabeled dataset.
The proposed method is two-fold; a style-extracting CVAE and a style-isolating CL model.
To extract styles using CVAEs effectively, we need proper conditions that are independent of styles.
To this end, we utilize the features of CL as conditions (soft labels) because CL models extract style-independent features~\cite{Kugelgen2021} and are trained without labels.
The CVAE and CL model form a parallel network in the proposed architecture, where the input data are fed into both, and the output of the CL model conditions the CVAE.
The training procedure of the proposed CVAE involves evaluating the mutual information (MI) of feature vectors arising from the CL model and the CVAE.
MI promotes the independence of the two vectors to enable the CVAE to extract only styles.
Experiments on two simple datasets, MNIST~\cite{MNIST} and an original dataset based on Google Fonts~\cite{googlefonts}, showed that the proposed method effectively extracts style features.
For practical results, we also evaluated the method using two natural image datasets, i.e.,\ Imagenette~\cite{imagenette} and DAISO-100~\cite{katoh2021dataset}.

%% file: relatedwork.tex
\section{Related Work}
DNNs can handle styles using style transfer tasks~\cite{Gatys2016,Jing2020}.
Given a content image and a style image, style-transferring DNNs generally generate images with the same content as the first input and with the same style as the second input.
They aim to render high-quality images in different styles.
Thus, the style-transfer methods do not assign weight to style features or embedding vectors of styles.

AEs~\cite{Vincent2008} can extract style features; in particular, styles in feature spaces can be observed using AEs that can bind feature spaces to specific distributions such as VAEs~\cite{Kingma2014} and adversarial AEs~\cite{Makhzani2016}.
AEs generally form clusters in feature spaces, and the styles are distributed in each cluster.
However, the styles are usually mixed with other content in the feature spaces.

Disentanglement~\cite{Higgins2017,Qi2022,locatello19a} is an approach to handle such mixed features.
Disentanglement methods attempt to separate the features or feature spaces into interpretable parts.
Several techniques, such as regularization and quantization, have been proposed to disentangle features.
However, extracting specific features using such methods is difficult~\cite{locatello19a}.

A more straightforward approach for extracting only styles is to use labels.
CVAEs~\cite{Kingma2014a,Sohn2015} use class labels as conditions and learn feature spaces that represent common styles across labels.
If labels are not available, the features or components selected for use as the condition for extracting only styles are not obvious.

CL methods, which are self-supervised representation learning methods, are potential choices for the condition.
CL methods find feature spaces robust against data augmentation, and thus they can be considered style-isolation methods~\cite{Kugelgen2021}.
CL itself is a popular representation learning method that has achieved great success, especially with image data~\cite{SimCLR,MoCo,SwAV,Le-Khac2020}.
SimCLR~\cite{SimCLR} showed that the framework of CL works well for images, and MoCo~\cite{MoCo} improved SimCLR's computation costs.
These methods use positive samples and many negative samples as inputs and draw a contrast between them to learn meaningful feature spaces.
Recently, CL methods without negative samples~\cite{SimSiam,VICReg} have been proposed.
These methods use techniques such as regularization and special network structures, and they are called \emph{noncontrastive} CL methods.

Styles can also be seen as domains.
Domain adaptation is a well-studied task that deals with changes in domains between the training and testing phases~\cite{Wang2018}.
Domain adaptation is achieved by obtaining domain-independent feature extractors.
To this end, various approaches have been proposed;
for example, evaluating domain independence by DNNs in an adversarial manner~\cite{Ganin2016},
regularizing the feature vector to have the same distribution across domains using the maximum mean discrepancy~\cite{Long2016},
and utilizing learnable classifiers for extracting classification-aware and domain-independent features~\cite{Saito2018}.
However, these domain adaptation methods aim to extract the common feature across different domains, and domain-specific features have not received considerable attention.

%% file: method.tex
\section{Method}
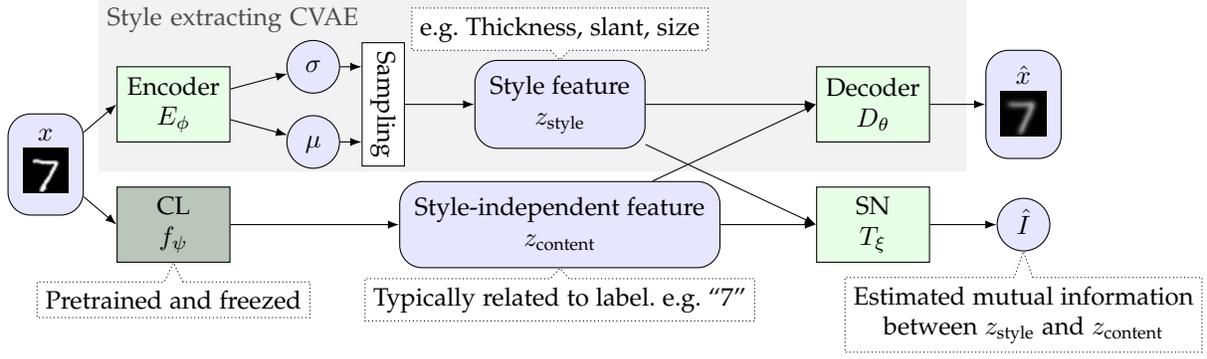
\begin{figure*}[t]
    \centering
    \begin{tikzpicture}[%
    data/.style={draw, circle, minimum height=0.7cm, inner sep=0cm, font=\vphantom{Ag}, fill=blue!10, align=center},%
    ldata/.style={data, rectangle, rounded corners=3mm, minimum width=1cm, inner sep=2mm},%
    op/.style={draw, rectangle, fill=white},%
    nnet/.style={op, minimum height=1cm, minimum width=1.5cm, fill=green!10, align=center},%
    flow/.style={-{Latex[length=1.5mm]}},%
    mycallout/.style={rectangle callout, draw, densely dotted, fill=white},%
    ]
        \node[fill=black!5, text=black!70, anchor=south west, align=left, text width=11.3cm] at (0.7, -0.1) {Style extracting CVAE\\[1.3cm]\strut};

        \node[ldata] (input) at (0, 0) {$x$\\[1mm]\includegraphics[width=0.6cm]{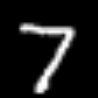}};
        \node[nnet] (enc_im) at (1.7, 0.8) {Encoder\\ $E_\phi$};
        \node[nnet, fill=green!10!gray!50] (enc_ex) at (1.7, -0.8) {CL\\ $f_\psi$};
        \node[data, right=1.5cm, yshift=-0.5cm] (mean) at (enc_im) {$\mu$};
        \node[data, right=1.5cm, yshift=0.5cm] (std) at (enc_im) {$\sigma$};
        \node[op, right=2.5cm] (sampling) at (enc_im){\rotatebox{270}{Sampling}};
        \node[ldata, right=4cm] (z_im) at (enc_im) {Style feature\\ $z_\text{style}$};
        \node[ldata, right=3cm] (z_ex) at (enc_ex) {Style-independent feature\\ $z_\text{content}$};
        \node[nnet] (dec) at (11, 0.8) {Decoder\\ $D_\theta$};
        \node[nnet] (sn) at (11, -0.8) {SN\\ $T_\xi$};
        \node[ldata] (output) at (13, 0.8) {$\hat{x}$\\[1mm]\includegraphics[width=0.6cm]{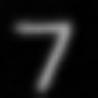}};
        \node[data] (mine) at (13, -0.8) {$\hat{I}$};

        \draw[flow] (input) -- (enc_im.west);
        \draw[flow] (input) -- (enc_ex.west);
        \draw[flow] (enc_im) -- (mean);
        \draw[flow] (enc_im) -- (std);
        \draw[flow] (mean) -- (mean-|sampling.west);
        \draw[flow] (std) -- (std-|sampling.west);
        \draw[flow] (sampling) -- (z_im);
        \draw[flow] (enc_ex) -- (z_ex);
        \draw[flow] (z_im) -- (dec);
        \draw[flow] (z_ex) -- (dec.west);
        \draw[flow] (z_im) -- (sn.west);
        \draw[flow] (z_ex) -- (sn);
        \draw[flow] (dec) -- (output);
        \draw[flow] (sn) -- (mine);

        \node[mycallout, callout absolute pointer=(enc_ex.south), yshift=-1cm] at (enc_ex){Pretrained and freezed};
        \node[mycallout, callout absolute pointer=(z_ex.south), yshift=-1cm] at (z_ex){Typically related to label. e.g.~``7''};
        \node[mycallout, callout absolute pointer=(z_im.north), yshift=1cm] at (z_im){e.g.~Thickness, slant, size};
        \node[mycallout, callout absolute pointer=(mine.south), yshift=-1.2cm, align=center] at (mine){Estimated mutual information\\ between $z_\text{style}$ and $z_\text{content}$};
    \end{tikzpicture}
    \caption{%
    Overview of the proposed method. The VAE conditioned by the CL model extracts the style features.
    In the training procedure, the estimated mutual information of the two feature vectors $z_\text{content}$ and $z_\text{style}$ is evaluated to encourage the features to be statistically independent.
    }
    \label{fig:overview}
\end{figure*}

In this section, we describe the proposed method for style feature extraction.
The proposed method includes a CVAE conditioned by CL and an MI estimator, as shown in \cref{fig:overview}.

We construct the CVAE for style extraction,
and introduce CL that extracts style-independent features as the condition of the CVAE.
Then, we present a constraint based on MI that helps CVAE extract only style information.
Finally, we explain the training procedure and loss function of the proposed method.

\subsection{Style Extracting CVAE}
Assume that data $X = \{x_1, x_2, \dots, x_N\}$ are generated based on style and other information.
Then, we consider a generative model
\begin{equation}
    x \sim p_\theta(x|z_\text{content}, z_\text{style}),
    \label{eq:cvae_generative_model}
\end{equation}
where $\theta, z_\text{content}$, and $z_\text{style}$ are the parameter of the distribution $p$, a style-independent variable, and a style-dependent variable, respectively.
The style-independent variable $z_\text{content}$ typically corresponds to class labels.

Given $z_\text{content}$ and assuming $z_\text{style}$ is drawn from standard Gaussian distribution, 
CVAE~\cite{Kingma2014a,Sohn2015} is used as a method for estimating $p_\theta$ conditioned by $z_\text{content}$.
In the framework of CVAE, we introduce an encoder $q_\phi(z_\text{style}|x)$ that estimates unobserved $z_\text{style}$, and then find $p_\theta, q_\phi$ that maximize the lower bound
\begin{align}
    \log p_\theta(x) \geq &-\text{KL}\left(q_\phi(z_\text{style}|x) \| p(z_\text{style})\right)\nonumber\\
    &+ \mathbb{E}_{q_\phi(z_\text{style}|x)}\left[\log p_\theta(x|z_\text{style}, z_\text{content})\right],
    \label{eq:elbo}
\end{align}
where $\text{KL}(\cdot\|\cdot)$ represents the Kullback--Leibler divergence (KLD) and $p(z_\text{style})$ represents standard Gaussian distribution.
In the context of AEs, $q_\phi$ corresponds to the encoder that extracts style features, and $q_\theta$ corresponds to the decoder that reconstructs input $x$ from feature vectors $z_\text{content}$ and $z_\text{style}$.
Note that $z_\text{content}$ is not fed to the encoder $q_\phi$ unlike in the typical setups of CVAE~\cite{Kingma2014a,Sohn2015} because our goal is to obtain an encoder that extracts style features from only the input data.

In practice, we utilize DNNs for implementing the encoder and decoder, which are denoted by $E_\phi$ and $D_\theta$, respectively.
Assuming that $p_\theta$ is an isotropic Gaussian distribution, the DNNs are trained using the empirical loss function given as
\begin{align}
    L_\text{CVAE}(\theta, \phi; X) =& \lambda_\text{KL} \frac{1}{2}\sum_{i=1}^{|X|}\left[1+\log(\sigma_i^2)-\mu_i^2-\sigma_i^2\right] \nonumber\\
    &\hspace{-1cm} + \lambda_\text{recon} \frac{1}{N}\sum_{x\in X} \|D_\theta(x; z_\text{style}, z_\text{content}) - x\|^2,
    \label{eq:loss_vae}
\end{align}
where $\lambda_\text{KL}$ and $\lambda_\text{recon}$ represent the weights of the KLD and reconstruction loss, respectively, and $\mu$ and $\sigma$ represent the output of encoder $E_\phi(x)$.

\subsection{CL as a Style-Independent Feature Extractor}
We introduce CL to obtain $z_\text{content}$ that we assumed given in \cref{eq:cvae_generative_model}.
CL~\cite{SimCLR,MoCo,SimSiam} is a self-supervised representation learning framework based on a simple idea: the two feature vectors should be the same if they are derived from the same data, but with different data augmentation operations.
CL models can be seen as style-independent feature extractors when considering data augmentation as a perturbation in styles~\cite{Kugelgen2021}.
Therefore, we can use pretrained CL models for the feature extractors that output $z_\text{content}$.

We briefly review a CL framework based on MoCo~\cite{MoCo}, which is popular and relatively lightweight.
In MoCo, a feature extractor $f_\psi$ with parameter $\psi$ is trained by minimizing InfoNCE given by
\begin{gather}
    L_\text{CL}(\psi) = -\log\frac{\exp(q\cdot k_+ / \tau)}{\exp(q\cdot k_+ / \tau) + \sum_{i=1}^K \exp(q\cdot k_- / \tau)},
    \label{eq:loss_cl}\\
    q = \text{MLPHead}(z_\text{content}), z_\text{content} = f_\psi(x),
\end{gather}
where $x, q, k_+, k_-$, and $K$ represent the input data, representation corresponding to $x$, positive key, negative key, and number of negative keys, respectively.
Note that $q, k_+$, and $k_-$ are normalized (i.e., $\|q\| = \|k_+\| = \|k_-\| = 1$).
The positive key $k_+$ is the representation corresponding to the same data as $x$ but with different data augmentation.
The negative key $k_-$ is the representation corresponding to the data different from $x$.
In general CL, we use the output of $\text{MLPHead}(\cdot)$ which consists of a few fully-connected layers to evaluate InfoNCE.
Once $f_\psi$ is trained, we employ $z_\text{content}$, which is the input of MLPHhead, for the inferred feature vector of the trained CL models.


\subsection{Mutual Information Constraint by MINE}
To ensure that the feature vector of the CVAE ($z_\text{style}$) only contains style information, we introduce a constraint to encourage the independence between $z_\text{content}$ and $z_\text{style}$.
The generative model presented in \cref{eq:cvae_generative_model} assumes that $z_\text{content}$ and $z_\text{style}$ are independent; however, this assumption does not always hold.
Specifically, the loss function of the CVAE given by \cref{eq:loss_vae} can be lowered by ignoring the condition $z_\text{content}$ when DNNs $E_\phi$ and $D_\theta$ have high degrees of freedom. 
In this case, $z_\text{style}$ has sufficient information for reconstructing the input.
This means that $z_\text{style}$ contains features other than styles, and $z_\text{style}$ and $z_\text{content}$ are not independent.
To alleviate this problem, we evaluate the independence of the two feature vectors.
MI is used as a measure to evaluate the independence between two variables~\cite{Bishop2006}.

By using the DNN-based MI estimator, MINE~\cite{MINE}, we measure the independence between $z_\text{content}$ and $z_\text{style}$.
In the MINE framework, we introduce a DNN $T_\xi$ called statistics network (SN) and determine $T_\xi$ that maximizes the lower bound of the MI given as
\begin{align}
    I(z_\text{content}; z_\text{style}) &\geq \hat{I}_\xi(z_\text{content}; z_\text{style}) \nonumber\\
    &\hspace{-2cm}= \mathbb{E}[T_\xi(z_\text{content}, z_\text{style})] - \log\left(\mathbb{E}[T_\xi(z_\text{content}, \bar{z}_\text{style})]\right),
    \label{eq:mine_kl}
\end{align}
where $\bar{z}_\text{style}$ represents a variable distributed the same as $z_\text{style}$.
We can draw $\bar{z}_\text{style}$ from standard Gaussian distribution because we assumed $z_\text{style}$ is standard Gaussian distributed in the CVAE.
The estimated MI is the maximized value of $\hat{I}_\xi$.
The actual learning procedure of SN tends to be unstable because the estimator given in \cref{eq:mine_kl} is based on asymmetric KLD~\cite{MINE,DeepInfomax}.
To alleviate this, we use a variant of the estimator that uses the Jensen--Shannon divergence instead of KLD~\cite{DeepInfomax};
\begin{align}
    \hat{I}_\xi^\text{JS}(z_\text{content}; z_\text{style})
    =\ & \mathbb{E}[-\text{sp}(-T_\xi(z_\text{content}, z_\text{style}))] \nonumber\\
    &- \mathbb{E}[\text{sp}(T_\xi(z_\text{content}, \bar{z}_\text{style}))],
    \label{eq:mine_jsd}
\end{align}
where $\text{sp}(\cdot)$ represents a softplus function~\cite{Softplus}.
Given observations $X$ and the corresponding feature vectors $Z_\text{content} = \{z_\text{content}^{(1)}, \dots, z_\text{content}^{(N)}\}$ and $Z_\text{style} = \{z_\text{style}^{(1)}, \dots, z_\text{style}^{(N)}\}$, the empirical form of \cref{eq:mine_jsd} becomes
\begin{align}
    \hat{I}_\xi^\text{Emp}(Z_\text{content}; Z_\text{style}) &=  \frac{1}{N}\sum_{n=1}^N -\text{sp}(-T_\xi(z_\text{content}^{(n)}, z_\text{style}^{(n)})) \nonumber\\
    &\hspace{2em}- \frac{1}{N}\sum_{n=1}^N\text{sp}(T_\xi(z_\text{content}^{(n)}, \bar{z}_\text{style}^{(n)})).
    \label{eq:emp_mine}
\end{align}
We ensure $z_\text{content}$ and $z_\text{style}$ are independent by adding this empirical MI estimator to the loss function of CVAE (\cref{eq:loss_vae}) in the training procedure.

\subsection{Training and Loss Function}
The training procedure of the proposed method consists of two steps.
As the first step, we train the CL model $f_\psi$ using any existing CL methods and freeze the parameters of the model.
The trained $f_\psi$ can be employed as a style-independent feature extractor.
In the second step, we train the CVAE for style feature extraction.
The training of CVAE is constrained by MINE to ensure that $z_\text{content}$ and $z_\text{style}$ are independent.
This constraint helps CVAE extract only style features.

The loss function for the encoder $E_\phi$ and the decoder $D_\theta$ is given as a linear combination of the CVAE loss (\cref{eq:loss_vae}) and the estimated MI (\cref{eq:emp_mine});
\begin{align}
    &L(\theta, \phi, \xi; X, Z_\text{content}, Z_\text{style}) \nonumber\\
    &= L_\text{CVAE}(\theta, \phi; X) + \lambda_\text{MINE} \hat{I}_\xi^\text{Emp}(Z_\text{content}; Z_\text{style})
\end{align}
where $\lambda_\text{MINE}$ represents the weight of the MI.
The estimated MI $\hat{I}_\xi^\text{Emp}$ needs to be maximized by training $T_\xi$.
Thus, the proposed model that consists of the CVAE and SN is trained using the min-max problem
\begin{equation}
    \min_{\theta, \phi} \max_{\xi} L(\theta, \phi, \xi; X, Z_\text{content}, Z_\text{style}).
\end{equation}
This problem is similar to that of adversarial learning (e.g., generative adversarial networks~\cite{GAN}) and can be optimized in the same manner.

%% file: experiments.tex
\section{Experiments}
We evaluated the effectiveness of the proposed style feature extraction method using four datasets.
We analyzed the validity of the proposed method using MNIST~\cite{MNIST}, an original Google Fonts-based dataset (hereinafter, Google Fonts dataset) because they are simple, small, and easy to handle.
In addition, we assessed the performance on practical, real-world natural images using Imagenette~\cite{imagenette}, and DAISO-100~\cite{katoh2021dataset}.

Although several CL methods are available, we mainly employed MoCo v2~\cite{chen2020mocov2}, which is relatively lightweight and stable.
In the experiments using MNIST and the Google Fonts dataset, we evaluated the combination of the proposed method and the different CL methods.

\subsection{Datasets}
\subsubsection{MNIST-Like Datasets}
\begin{figure}
    \centering
    \includegraphics[width=\linewidth]{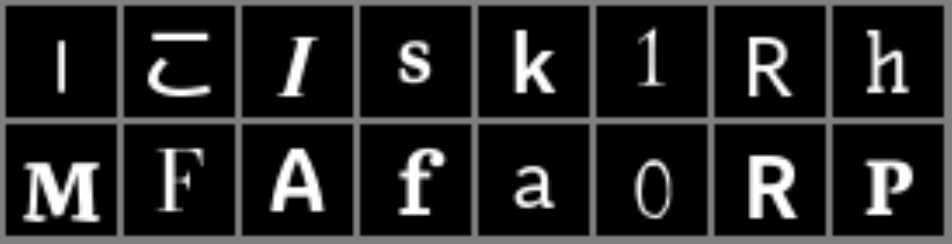}
    \caption{Example images from the Google Fonts dataset.}
    \label{fig:google_fonts_example}
\end{figure}

We confirmed that the proposed method works as intended using MNIST~\cite{MNIST} and the Google Fonts dataset.
MNIST is a handwritten digits image dataset that has various handwriting styles.
For additional evaluation, we composed the Google Fonts dataset, which is based on Google Fonts~\cite{googlefonts}.
Font faces can be considered as styles, and therefore, the Google Fonts dataset is expected to contain more style variations than that in MNIST.
The Google Fonts dataset consists of images of digits, Latin alphabets, and Japanese hiragana.
We obtained 1,244 font families by selecting fonts that contain these characters by creating images with each character for each font.
Each font family may contain different font weights.
We obtained 280,694 training samples and 31,189 testing samples.
Each image is $32\times 32$ pixels in grayscale.
Some examples from the Google Fonts dataset are shown in \cref{fig:google_fonts_example}.

For experiments using MNIST, we trained the CL model using EMNIST ByClass~\cite{emnist}, which is an extension of MNIST.
This is because CL works better with larger datasets~\cite{SimCLR}.
We trained the CVAE using MNIST and evaluated the overall proposed method also using MNIST.

We evaluated the proposed method from five viewpoints using the two datasets.
First, we confirmed that the CL method extracts feature vectors independent of data augmentation, and the CVAE successfully extracts features corresponding to the augmentation by observing the decoder outputs corresponding to the data augmented inputs.
Second, we assessed the relationship between the feature vectors of the CVAE and the styles by generating samples using the decoder with some fixed $z_\text{content}$ and interpolated $z_\text{style}$.
Third, we evaluated the variations in style features through style transfer experiments.
Fourth, we observed the neighbors of the test data mapped by the encoder (i.e., the test data in the space of $z_\text{style}$) to evaluate the style features without passing them to the decoder.
Finally, we compared the effects of using CL methods besides MoCo v2 in combination with the proposed method.

We used almost the same hyperparameters for MNIST and the Google Fonts dataset.
The backbone networks of the CL ($f_\psi$) and the encoder ($E_\phi$) were ResNet-18~\cite{He2016}.
While training the CL and the CVAE, data augmentation that consists of random perspective transformation, random cropping, random blurring, and random perturbation of brightness and contrast, was applied.
We used the Adam optimizer~\cite{Adam} for training.
Other hyperparameters are listed in \cref{tab:hparams}.
\begin{table*}[t]
    \centering
    \caption{Hyperparameters in the experiments.}
    \label{tab:hparams}
    \begin{tabular}{lccccc}
        \toprule
        Dataset      & Architecture of $E_\theta, f_\psi$ & Dataset for CL & $\lambda_\text{KL}$ & $\lambda_\text{MINE}$ & $\text{dim}(z_\text{style})$ \\\midrule
        MNIST        & ResNet-18 & EMNIST~\cite{emnist} ByClass & $0.1$ & $10^{-2}$ & 32\\
        Google Fonts & ResNet-18 & Google Fonts                 & $0.1$ & $10^{-3}$ & 32\\
        Imagenette   & ResNet-50 & ImageNet~\cite{ILSVRC15}     & $1$   & $10^{-2}$ & 128\\
        DAISO-100    & ResNet-50 & ImageNet~\cite{ILSVRC15}     & $1$   & $10^{-3}$ & 128\\\bottomrule
    \end{tabular}
\end{table*}

\subsubsection{Real-World Natural Image Datasets}
We conducted additional experiments using two natural image datasets, Imagenette~\cite{imagenette} and DAISO-100~\cite{katoh2021dataset}, to evaluate the expandability of the proposed method.
Imagenette is a lightweight subset of ImageNet~\cite{ILSVRC15} and contains 10 classes of the original ImageNet.
DAISO-100 is an image dataset consisting of photos of 100 miscellaneous goods in various scenarios.
In DAISO-100, three style-like conditions are explicitly labeled; these conditions are lighting, decoration by sticker, and camera angles.
We used ImageNet for training the CL model in the experiments using the natural image datasets.

We performed evaluations using style transfer and neighbor analysis because the proposed method does not aim to generate images.
The generated images were not sufficiently clear to be evaluated.
Style transfer experiments actually require image generation; however, we could examine changes in generated images when the styles were transferred.
We can evaluate the style features without generating images using neighbor analysis.

For the CL model ($f_\psi$) and encoder ($E_\phi$), we used ResNet-50~\cite{He2016}.
We initialized the CL model by pretrained weights, which are available publicly~\cite{chen2020mocov2}.
While training the CL model and CVAE, data augmentation used in MoCo v2~\cite{chen2020mocov2} was applied.
We used the Adam optimizer~\cite{Adam} for the training.
Other hyperparameters are summarized in \cref{tab:hparams}.

\subsection{Capturing Isolated Data Augmentation Features}
\label{sec:da_removal_mnist}
We conducted simple sanity check experiments.
For CL methods that learn features independent of data augmentation, we comfirmed that the CVAE successfully captured the information corresponding to data augmentation.

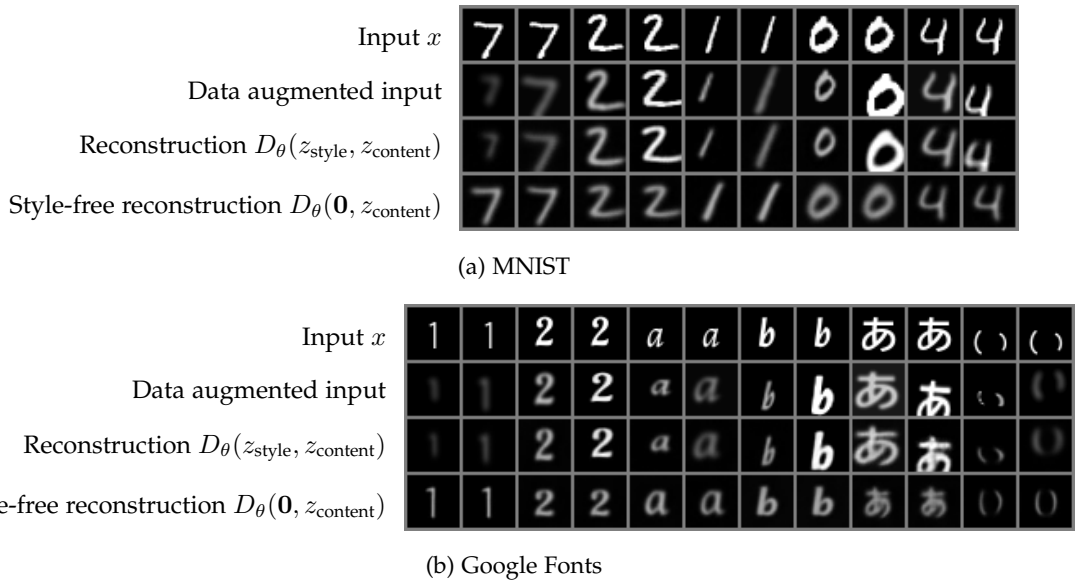
\begin{figure*}[t]
    \centering
    \begin{minipage}{\linewidth}
        \centering
        \begin{tikzpicture}
            \node (pic) at (0, 0) {\includegraphics[height=3cm]{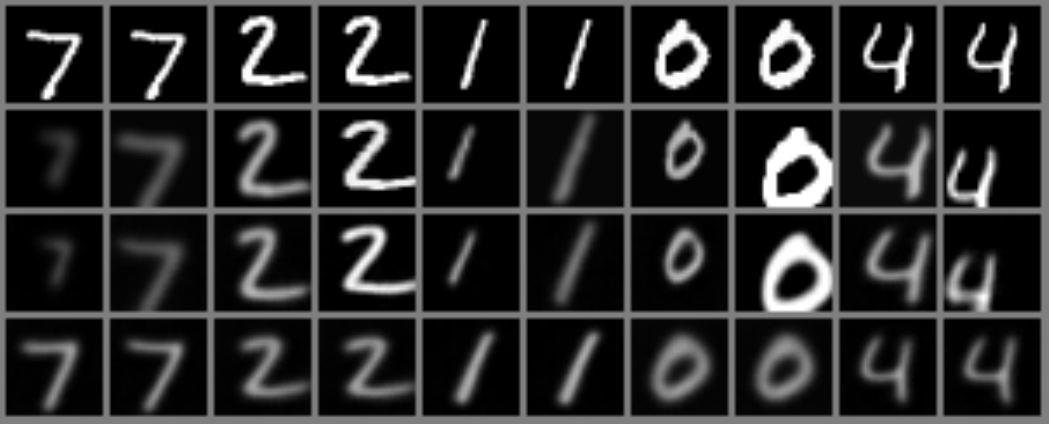}};
            \node[anchor=north east, yshift=-0.3cm] at (pic.north west) {Input $x$};
            \node[anchor=north east, yshift=-1cm] at (pic.north west) {Data augmented input};
            \node[anchor=north east, yshift=-1.7cm] at (pic.north west) {Reconstruction $D_\theta(z_\text{style}, z_\text{content})$};
            \node[anchor=north east, yshift=-2.5cm] at (pic.north west) {Style-free reconstruction $D_\theta(\mathbf{0}, z_\text{content})$};
        \end{tikzpicture}
        \subcaption{MNIST}\label{fig:da_removal_mnist}
    \end{minipage}
    \vspace{0.2cm}

    \begin{minipage}{\linewidth}
        \centering
        \begin{tikzpicture}
            \node (pic) at (0, 0) {\includegraphics[height=3cm]{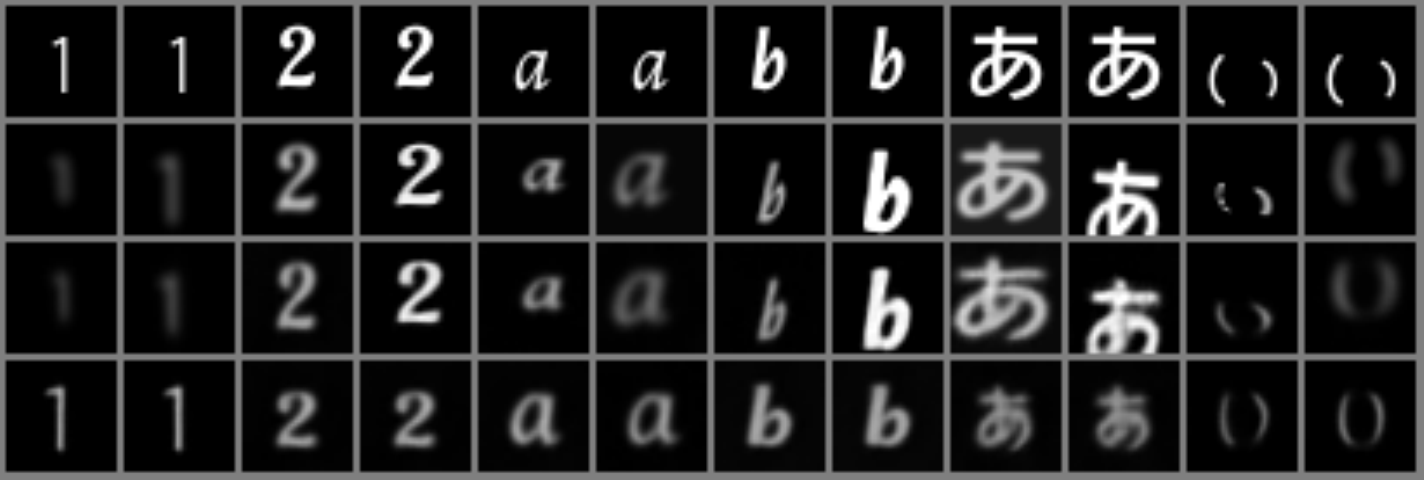}};
            \node[anchor=north east, yshift=-0.3cm] at (pic.north west) {Input $x$};
            \node[anchor=north east, yshift=-1cm] at (pic.north west) {Data augmented input};
            \node[anchor=north east, yshift=-1.7cm] at (pic.north west) {Reconstruction $D_\theta(z_\text{style}, z_\text{content})$};
            \node[anchor=north east, yshift=-2.5cm] at (pic.north west) {Style-free reconstruction $D_\theta(\mathbf{0}, z_\text{content})$};
        \end{tikzpicture}
        \subcaption{Google Fonts}\label{fig:da_removal_font}
    \end{minipage}
    \caption{%
    Examples of isolating and capturing data augmentation features.
    From top to bottom: original images, images after data augmentation, reconstructed images $D_\theta(z_\text{style}, z_\text{content})$, and images reconstructed without style features $D_\theta(\mathbf{0}, z_\text{content})$.
    }
    \label{fig:da_removal}
\end{figure*}

\Cref{fig:da_removal} shows the results.
In the figure, two types of reconstruction are shown: normal reconstruction using both $z_\text{content}$ and $z_\text{style}$ (the third row of \cref{fig:da_removal}) and data augmentation-independent reconstruction using only $z_\text{content}$ (the bottom row of \cref{fig:da_removal}).
Although normal reconstructions were similar to the input, the data augmentation-independent reconstructions looked similar to the input without augmentation effects.
These results illustrate that the CL model isolated data augmentation features and the CVAE captured them.
Interestingly, for the Google Fonts dataset, the reconstructions with $z_\text{style}=\mathbf{0}$ were not similar to the input data before the data augmentation; they seemed to be average style versions of the data-augmented input data.

\subsection{Conditional Generation}
\label{sec:interp_mnist}
Conditional generation experiments confirmed that the learned feature of the CVAE corresponded to the style feature.
We observed the decoder outputs with fixed $z_\text{content}$ and interpolated $z_\text{style}$.
The CL features $z_\text{content}$ were outputs of the CL model of the test data.
The CVAE features $z_\text{style}$ were generated along random line segments that cross the origin.

\begin{figure*}[t]
    \centering
    \begin{minipage}{0.5\linewidth}
        \centering
        \begin{tikzpicture}
            \node (pic) at (0, 0) {\includegraphics[width=0.85\linewidth]{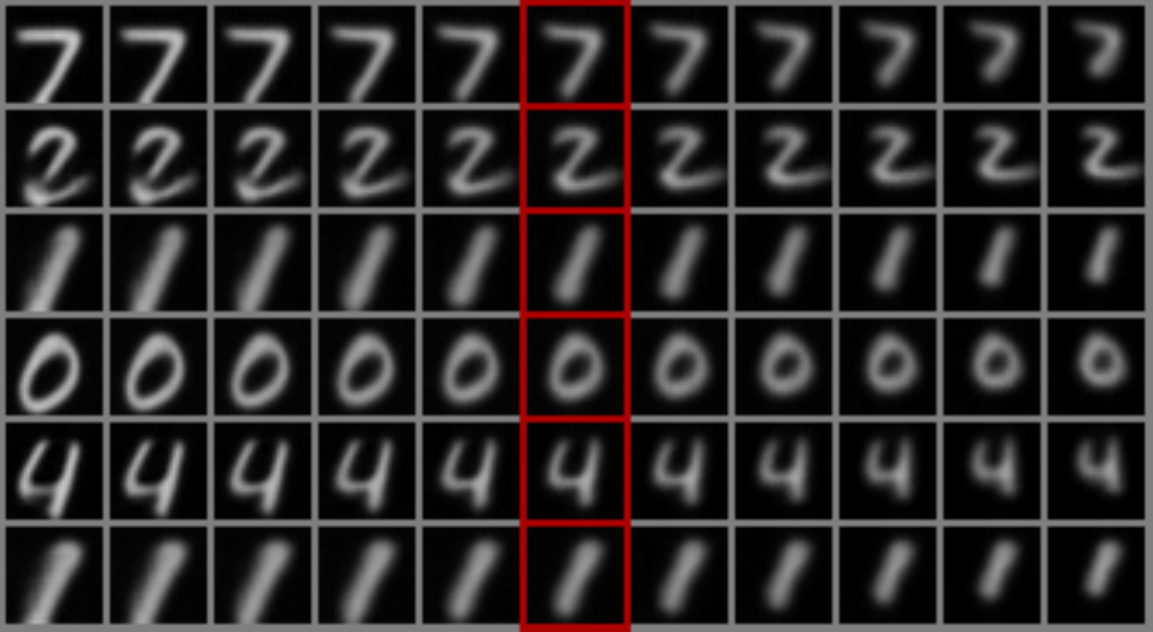}};
            \node[anchor=south east, xshift=1cm] at (pic.north west) {$z_\text{content}$};
            \node[anchor=north west, xshift=-0.8cm] at (pic.south east) {$z_\text{style}$};
            \draw[-latex](pic.south west) -- (pic.north west);
            \draw[-latex](pic.south west) -- (pic.south east);
        \end{tikzpicture}
        \subcaption{MNIST}\label{fig:interp_mnist}
    \end{minipage}%
    \begin{minipage}{0.5\linewidth}
        \centering
        \begin{tikzpicture}
            \node (pic) at (0, 0) {\includegraphics[width=0.85\linewidth]{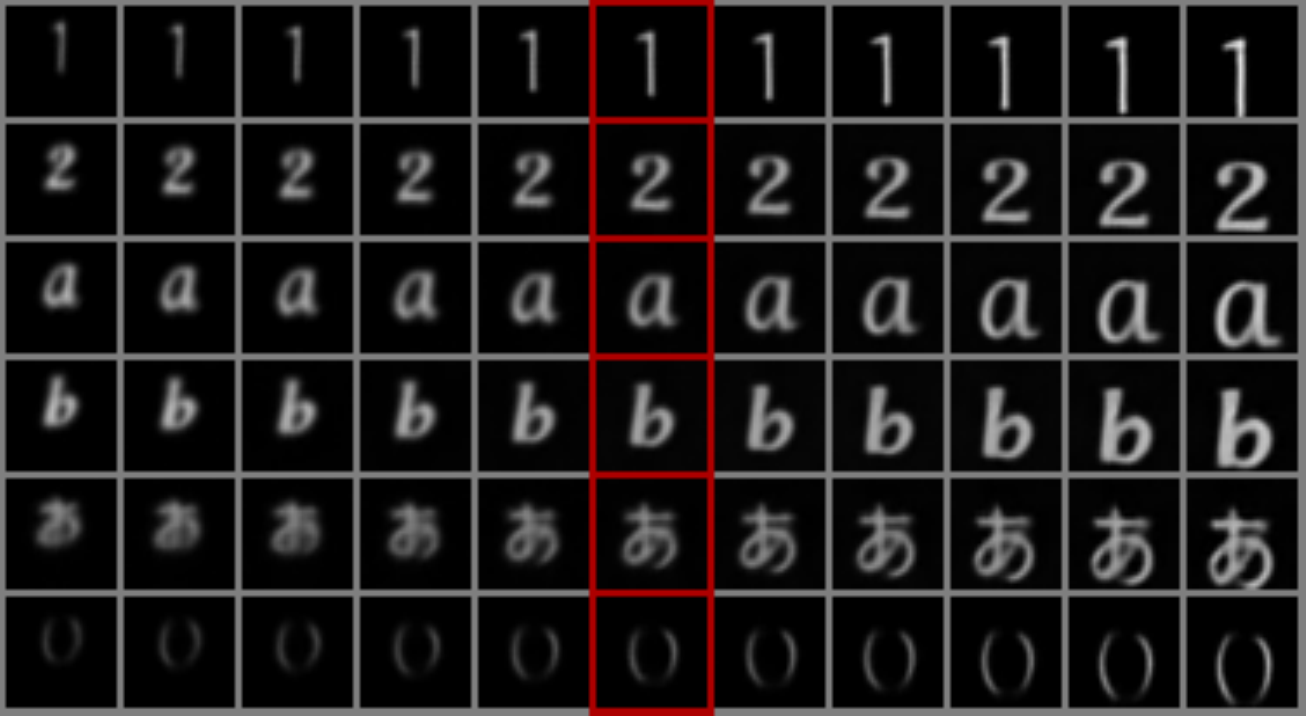}};
            \node[anchor=south east, xshift=1cm] at (pic.north west) {$z_\text{content}$};
            \node[anchor=north west, xshift=-0.8cm] at (pic.south east) {$z_\text{style}$};
            \draw[-latex](pic.south west) -- (pic.north west);
            \draw[-latex](pic.south west) -- (pic.south east);
        \end{tikzpicture}
        \subcaption{Google Fonts}\label{fig:interp_font}
    \end{minipage}
    \caption{%
    Examples of conditional generation.
    Each row corresponds to different $z_\text{content}$ from some test data.
    Each column corresponds to interpolated $z_\text{style} (0 \leq \|z_\text{style}\| \leq 3)$ along a randomly chosen unit vector.
    The images of the center column with the red frame are generated without the style features (i.e., $z_\text{style}=\mathbf{0}$).
    }
    \label{fig:interp}
\end{figure*}
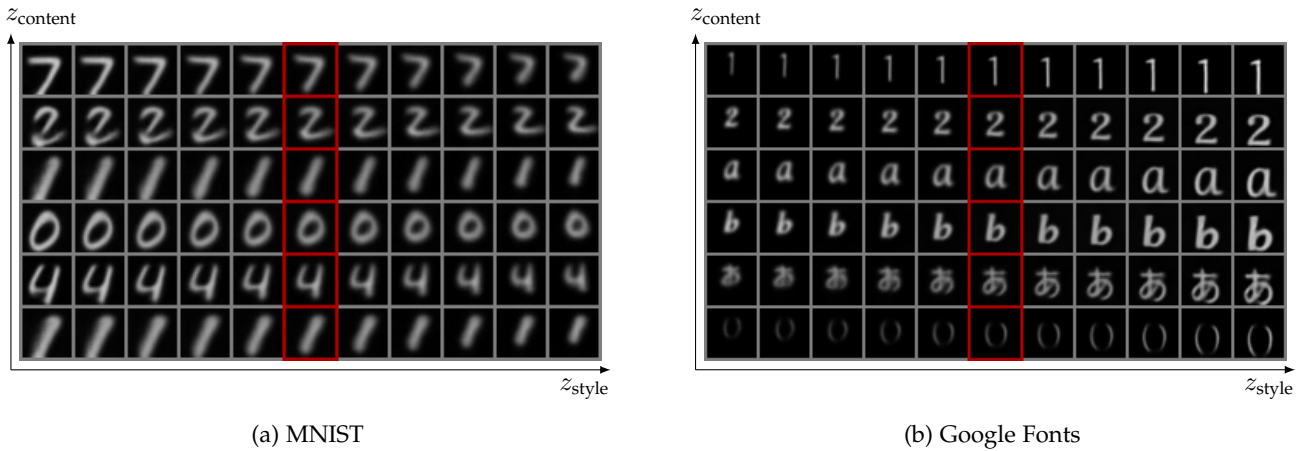

\Cref{fig:interp} shows the conditional generation results.
For both datasets, the styles changed by changing $z_\text{style}$ without altering the contents.
When $z_\text{style}$ was near the origin, the outputs were in average styles, and only styles shifted gradually as $z_\text{style}$ moved away from the origin.
In addition, changes in styles were consistent across different $z_\text{content}$.

\subsection{Style Transfer}
\label{sec:style_transfer_mnist}
Style-transfer experiments were performed to evaluate the variation of the learned style features.
Using the CL model, we extracted the $z_\text{content}$ of the test data.
Then, we combined them with the $z_\text{style}$ of different test data and put them into the decoder to generate style-transferred images.

\begin{figure*}[t]
    \centering
    \begin{minipage}[b]{0.45\linewidth}
        \centering
        \begin{tikzpicture}
            \node (pic) at (0, 0) {\includegraphics[height=6cm]{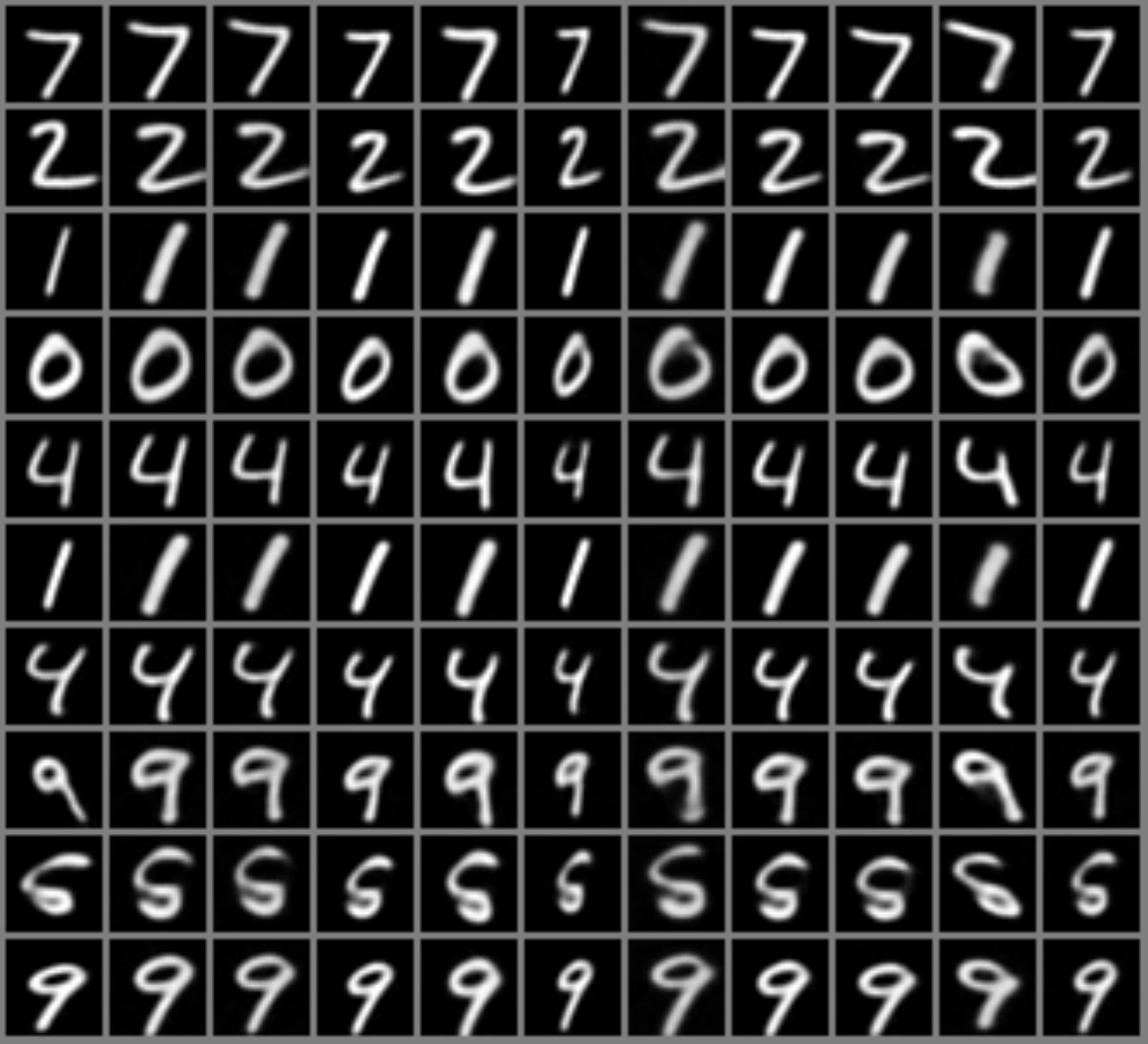}};
            \node[anchor=east] (src) at (pic.west) {\includegraphics[height=6cm]{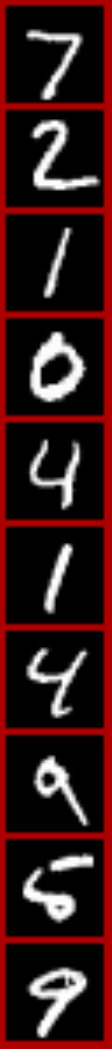}};
            \node[anchor=south] (dst) at (pic.north) {\includegraphics[width=6.6cm]{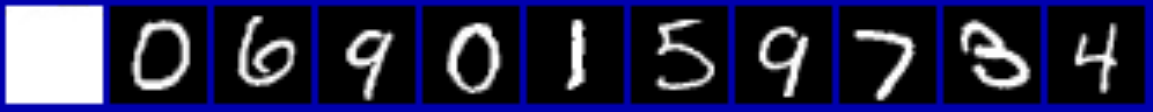}};
            \node[anchor=west, font=\footnotesize, xshift=0.08cm] at (dst.west) {self};
            \node[rotate=90, text=sxfer-red, anchor=south] at (src.west){Content source};
            \node[text=sxfer-blue, anchor=south] at (dst.north){Style destination};
        \end{tikzpicture}
        \subcaption{MNIST}\label{fig:style_transfer_mnist}
    \end{minipage}%
    \begin{minipage}[b]{0.55\linewidth}
        \centering
        \begin{tikzpicture}
            \node (pic) at (0, 0) {\includegraphics[height=6cm]{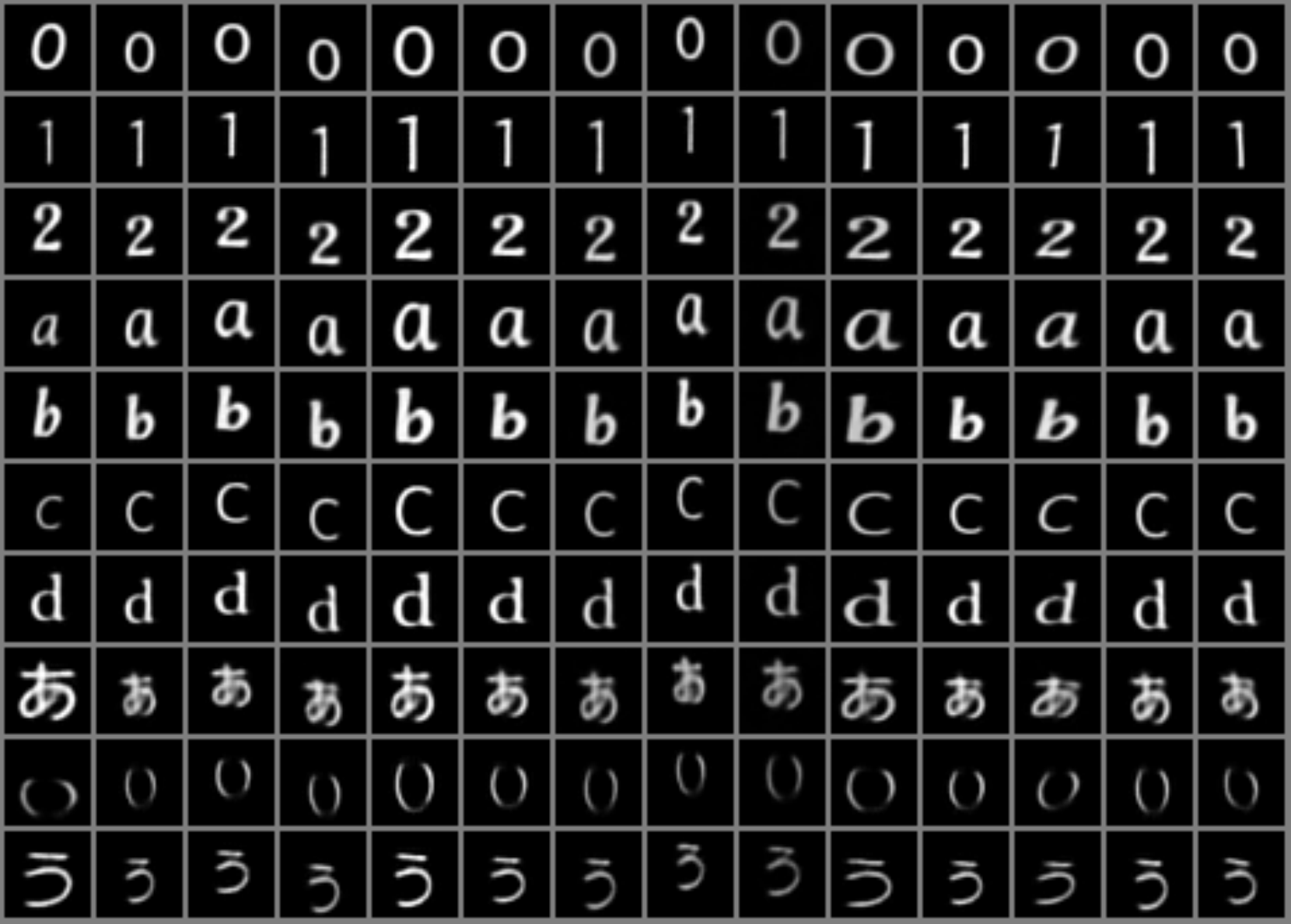}};
            \node[anchor=east] (src) at (pic.west) {\includegraphics[height=6cm]{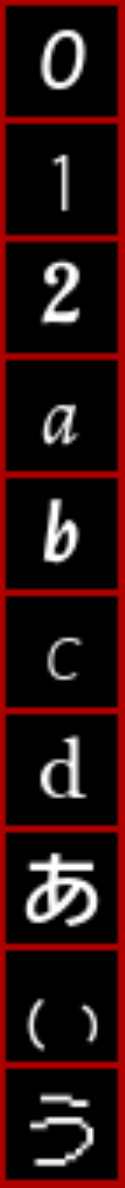}};
            \node[anchor=south] (dst) at (pic.north) {\includegraphics[width=8.4cm]{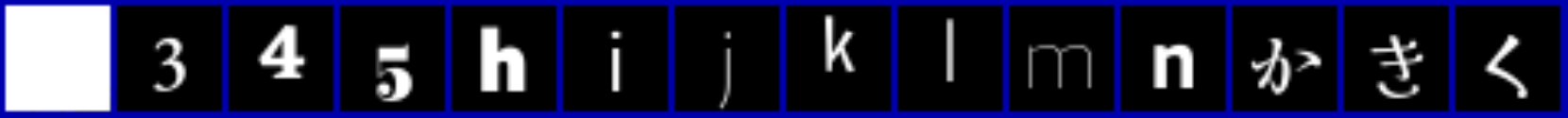}};
            \node[anchor=west, font=\footnotesize, xshift=0.08cm] at (dst.west) {self};
        \end{tikzpicture}
        \subcaption{Google Fonts}\label{fig:style_transfer_font}
    \end{minipage}
    \vspace{0.2cm}

    \begin{minipage}[b]{0.5\linewidth}
        \centering
        \begin{tikzpicture}
            \node (pic) at (0, 0) {\includegraphics[height=4cm]{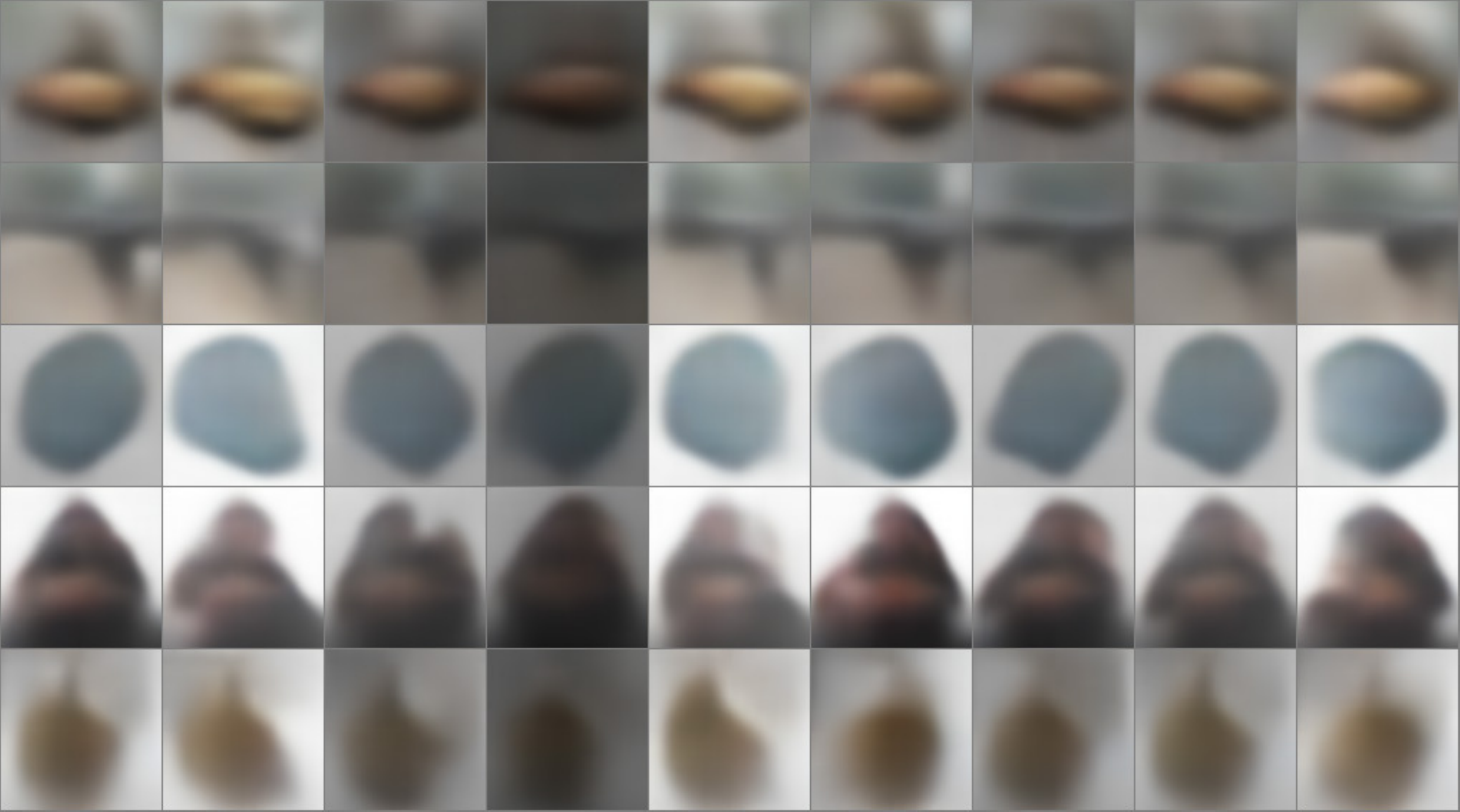}};
            \node[anchor=east] (src) at (pic.west) {\includegraphics[height=4cm]{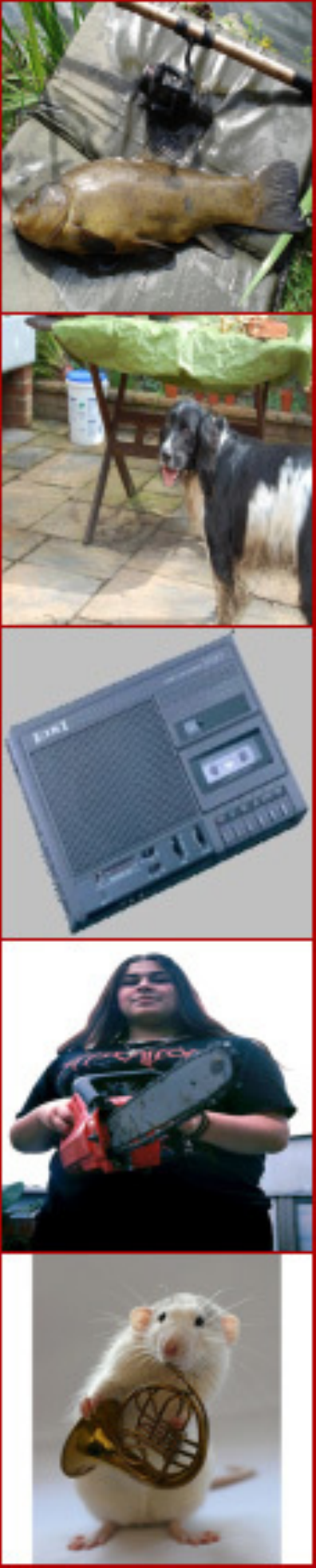}};
            \node[anchor=south] (dst) at (pic.north) {\includegraphics[width=7.2cm]{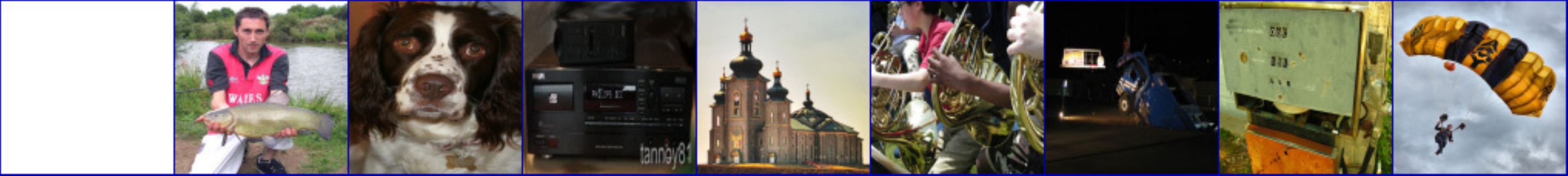}};
            \node[anchor=west, font=\footnotesize, xshift=0.2cm] at (dst.west) {self};
        \end{tikzpicture}
        \subcaption{Imagenette}\label{fig:style_transfer_imagenette}
    \end{minipage}%
    \begin{minipage}[b]{0.5\linewidth}
        \centering
        \begin{tikzpicture}
            \node (pic) at (0, 0) {\includegraphics[height=4cm]{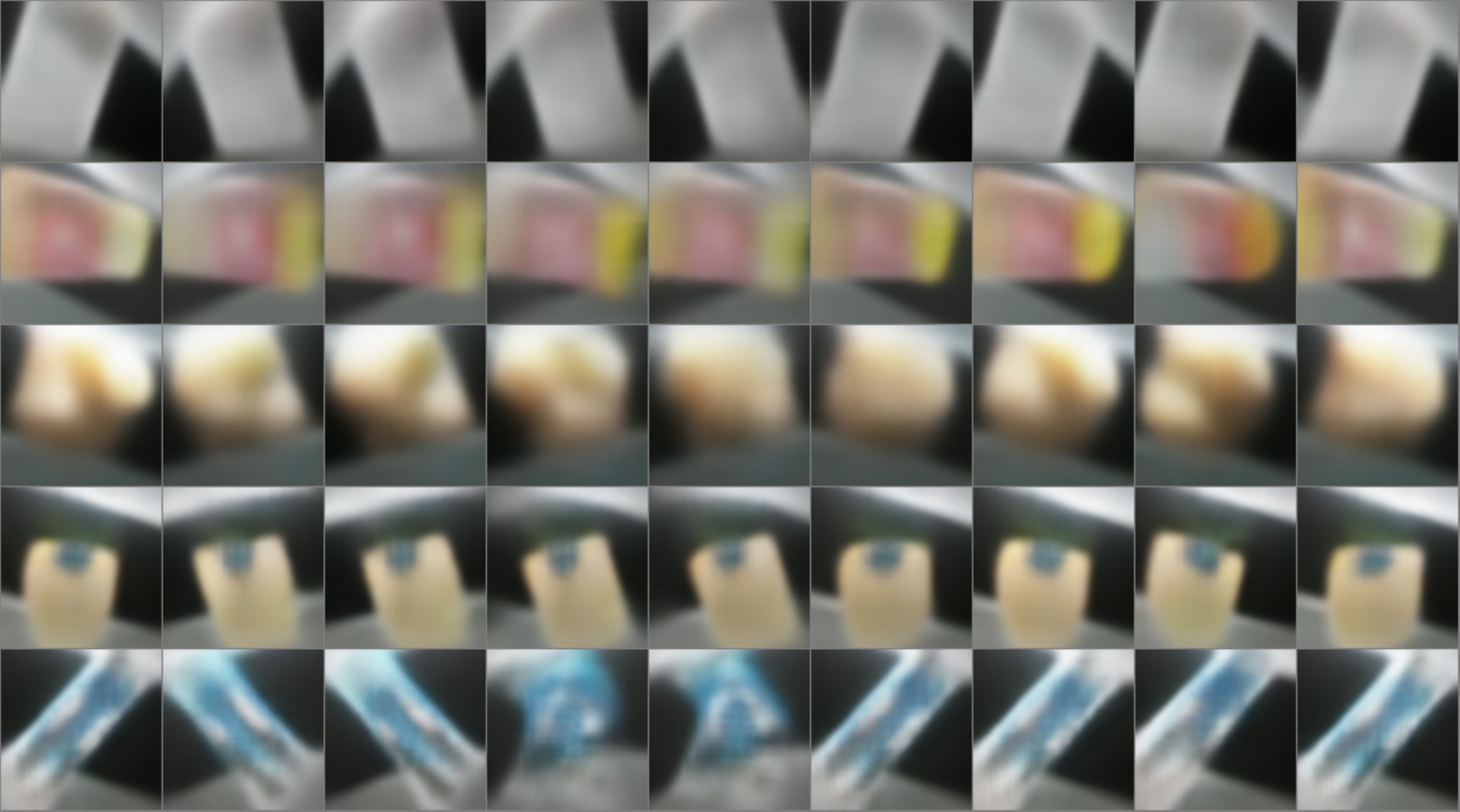}};
            \node[anchor=east] (src) at (pic.west) {\includegraphics[height=4cm]{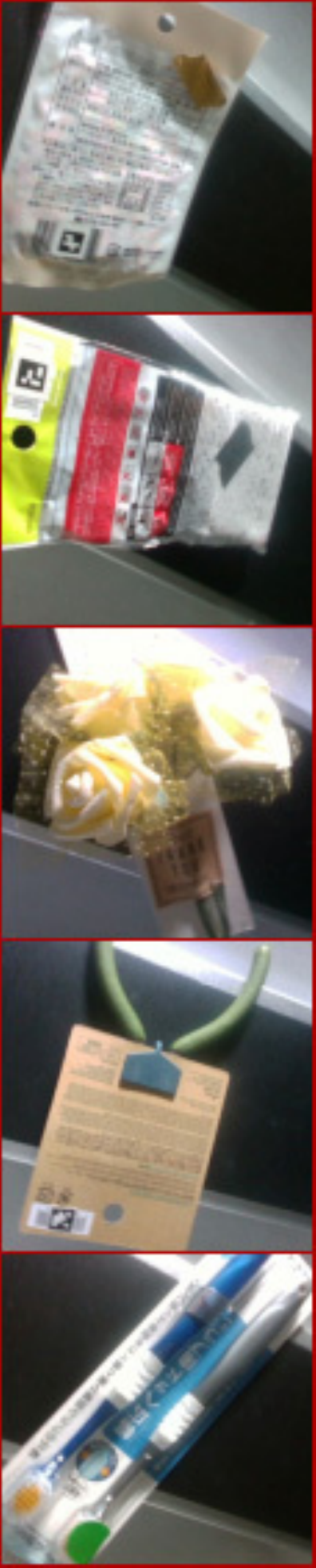}};
            \node[anchor=south] (dst) at (pic.north) {\includegraphics[width=7.2cm]{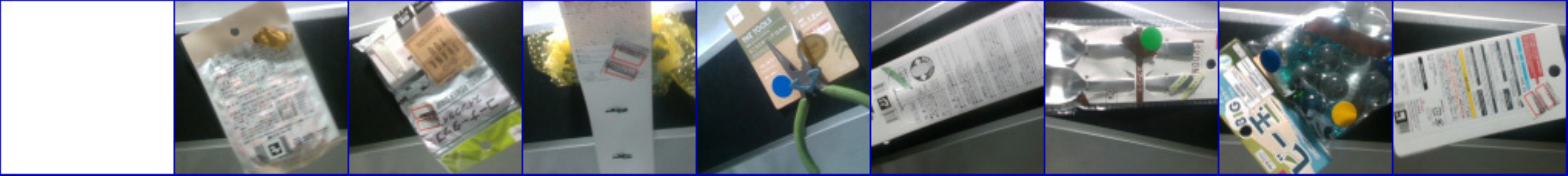}};
            \node[anchor=west, font=\footnotesize, xshift=0.2cm] at (dst.west) {self};
        \end{tikzpicture}
        \subcaption{DAISO-100}\label{fig:style_transfer_daiso100}
    \end{minipage}
    \caption{%
    Examples of style transfer.
    The red framed images represent the original images and the blue framed images represent the style destination images.
    Each white framed image is generated using the style-independent feature $z_\text{content}$ from the red framed image and the style feature $z_\text{style}$ from the blue framed image.
    }
    \label{fig:style_transfer}
\end{figure*}

The results of the style-transfer experiments are presented in \cref{fig:style_transfer}.
They illustrate that the styles were successfully transferred using the proposed style feature extractor.
For MNIST, styles such as the slant, size, and thickness were transferred.
For the Google Fonts dataset, mainly the bounding boxes (i.e., the size and the placement) of the characters were transferred as the styles.
This result indicated that the proposed method did not always learn the font faces as styles.

\cref{fig:style_transfer_imagenette} shows the results on Imagenette.
The generated images were not clear; however, styles such as brightness and object shapes were transferred.
The results on DAISO-100 are shown in \cref{fig:style_transfer_daiso100}.
The generated images were slightly clearer than those of Imagenette, and the direction of the background conveyor belt was transferred as styles.

\subsection{Neighbor Analysis}
\label{sec:nn_mnist}
Although we evaluated the outputs of the decoder in the previous three experiments, we also attempted to directly observe the style features by analyzing neighbors.
We mapped all test data to the style feature space using the encoder, and we observed the neighbors of some of the test data in terms of $z_\text{style}$.
Besides, we observed the neighbors in terms of $z_\text{content}$ to confirm the style independence of the CL features.

\begin{figure*}[t]
    \centering
    \begin{minipage}{0.5\linewidth}
        \centering
        \begin{tikzpicture}
            \node (pic) at (0, 0){\includegraphics[height=4.5cm]{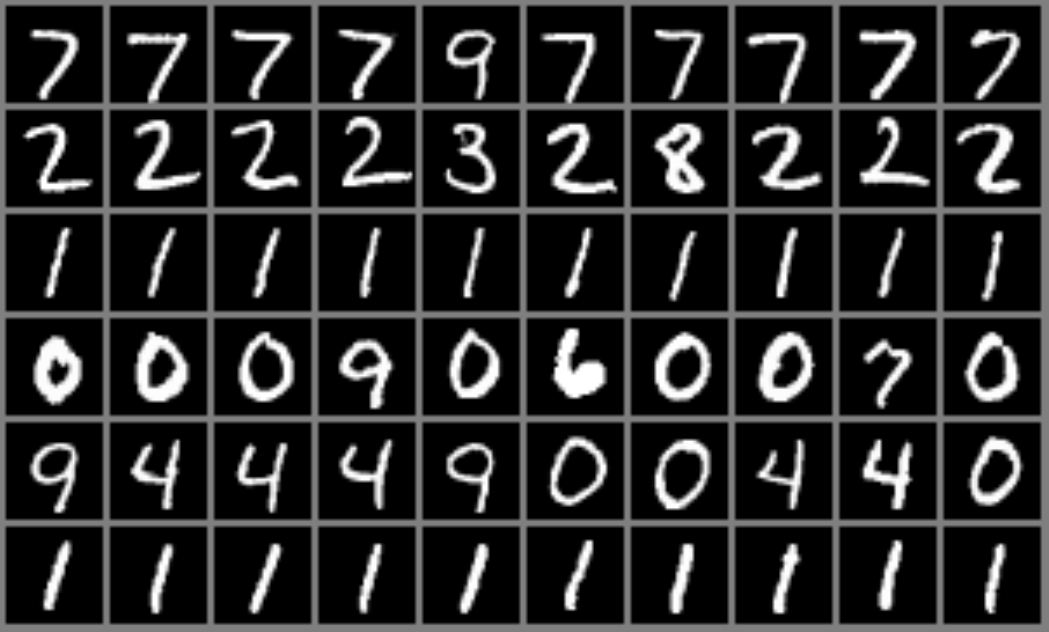}};
            \node[anchor=east] (anchors) at (pic.west){\includegraphics[height=4.5cm]{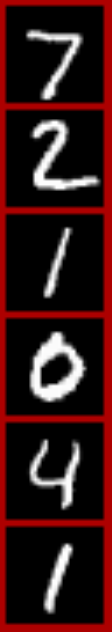}};
            \node[anchor=south, text=sxfer-red, font=\small] at (anchors.north) {Anchor};
            \node[anchor=south, font=\small, text depth=0cm] at (pic.north) {Neighbor data (w.r.t.~$z_\text{style}$)};
            \draw[-latex] (pic.south west) -- (pic.south east);
            \node[anchor=north east, font=\small] at (pic.south east) {Distance from anchors};
        \end{tikzpicture}
        \subcaption{MNIST, $z_\text{style}$}\label{fig:nn_vae_mnist}
    \end{minipage}%
    \begin{minipage}{0.5\linewidth}
        \centering
        \begin{tikzpicture}
            \node (pic) at (0, 0){\includegraphics[height=4.5cm]{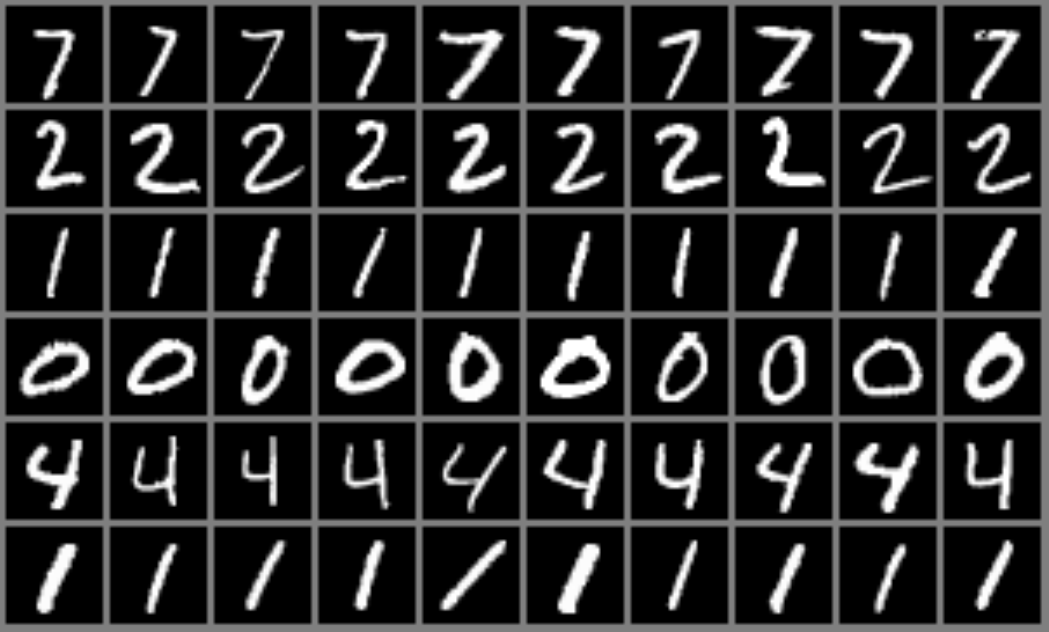}};
            \node[anchor=east] (anchors) at (pic.west){\includegraphics[height=4.5cm]{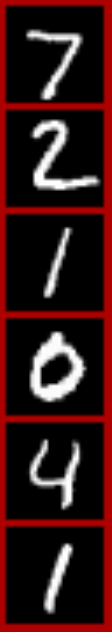}};
            \node[anchor=south, text=sxfer-red, font=\small] at (anchors.north) {Anchor};
            \node[anchor=south, font=\small, text depth=0cm] at (pic.north) {Neighbor data (w.r.t.~$z_\text{content}$)};
            \draw[-latex] (pic.south west) -- (pic.south east);
            \node[anchor=north east, font=\small] at (pic.south east) {Distance from anchors};
        \end{tikzpicture}
        \subcaption{MNIST, $z_\text{content}$}\label{fig:nn_cl_mnist}
    \end{minipage}
    \vspace{0.2cm}

    \begin{minipage}{0.5\linewidth}
        \centering
        \begin{tikzpicture}
            \node (pic) at (0, 0){\includegraphics[height=4.5cm]{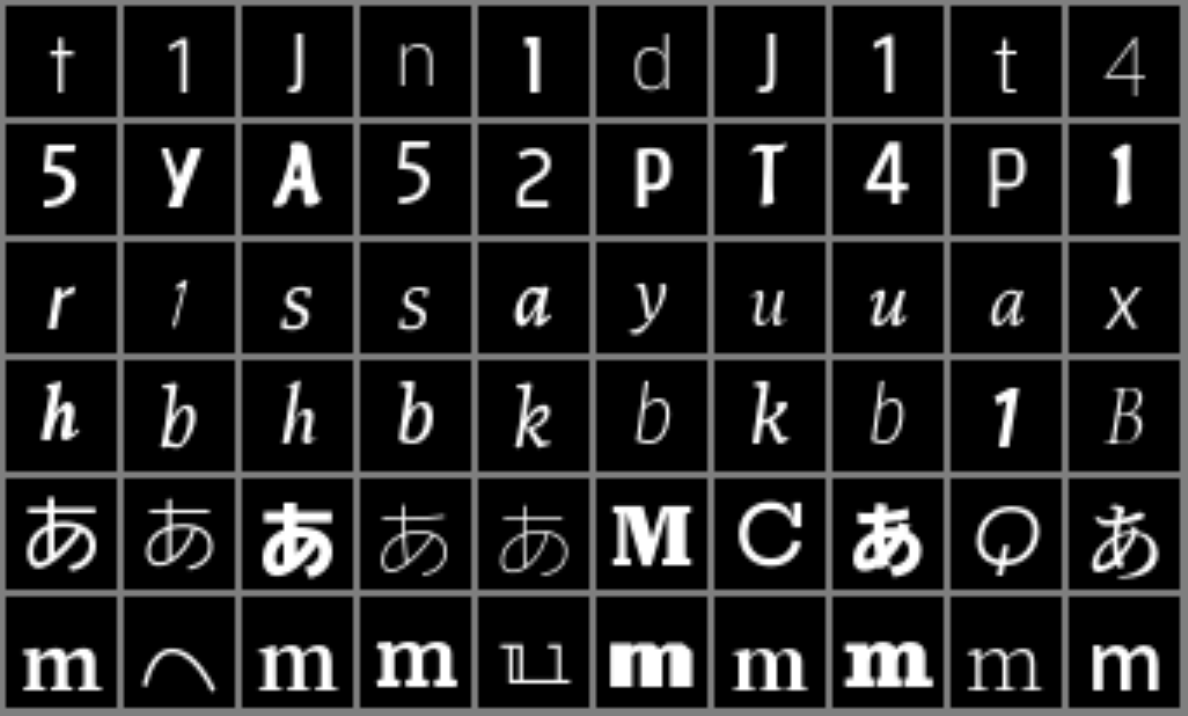}};
            \node[anchor=east] (anchors) at (pic.west){\includegraphics[height=4.5cm]{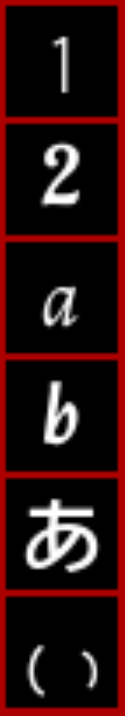}};
        \end{tikzpicture}
        \subcaption{Google Fonts, $z_\text{style}$}\label{fig:nn_vae_font}
    \end{minipage}%
    \begin{minipage}{0.5\linewidth}
        \centering
        \begin{tikzpicture}
            \node (pic) at (0, 0){\includegraphics[height=4.5cm]{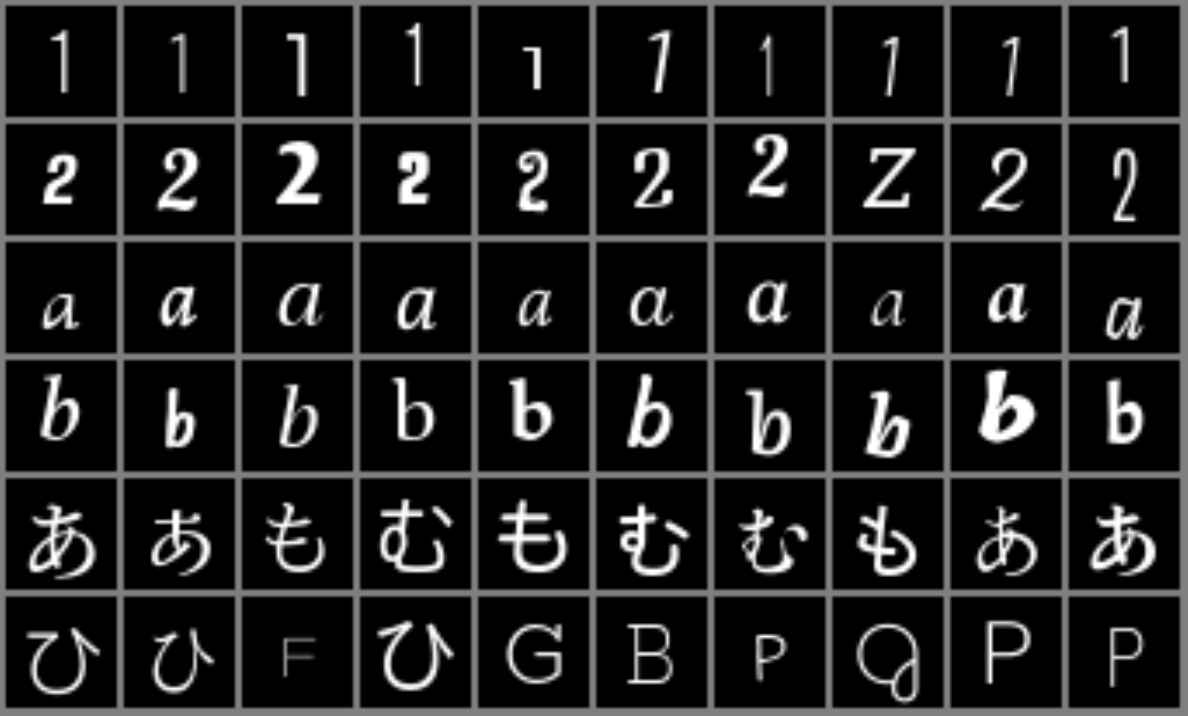}};
            \node[anchor=east] (anchors) at (pic.west){\includegraphics[height=4.5cm]{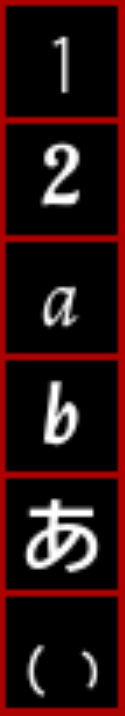}};
        \end{tikzpicture}
        \subcaption{Google Fonts, $z_\text{content}$}\label{fig:nn_cl_font}
    \end{minipage}
    \vspace{0.2cm}

    \begin{minipage}{0.5\linewidth}
        \centering
        \begin{tikzpicture}
            \node (pic) at (0, 0){\includegraphics[height=3.8cm]{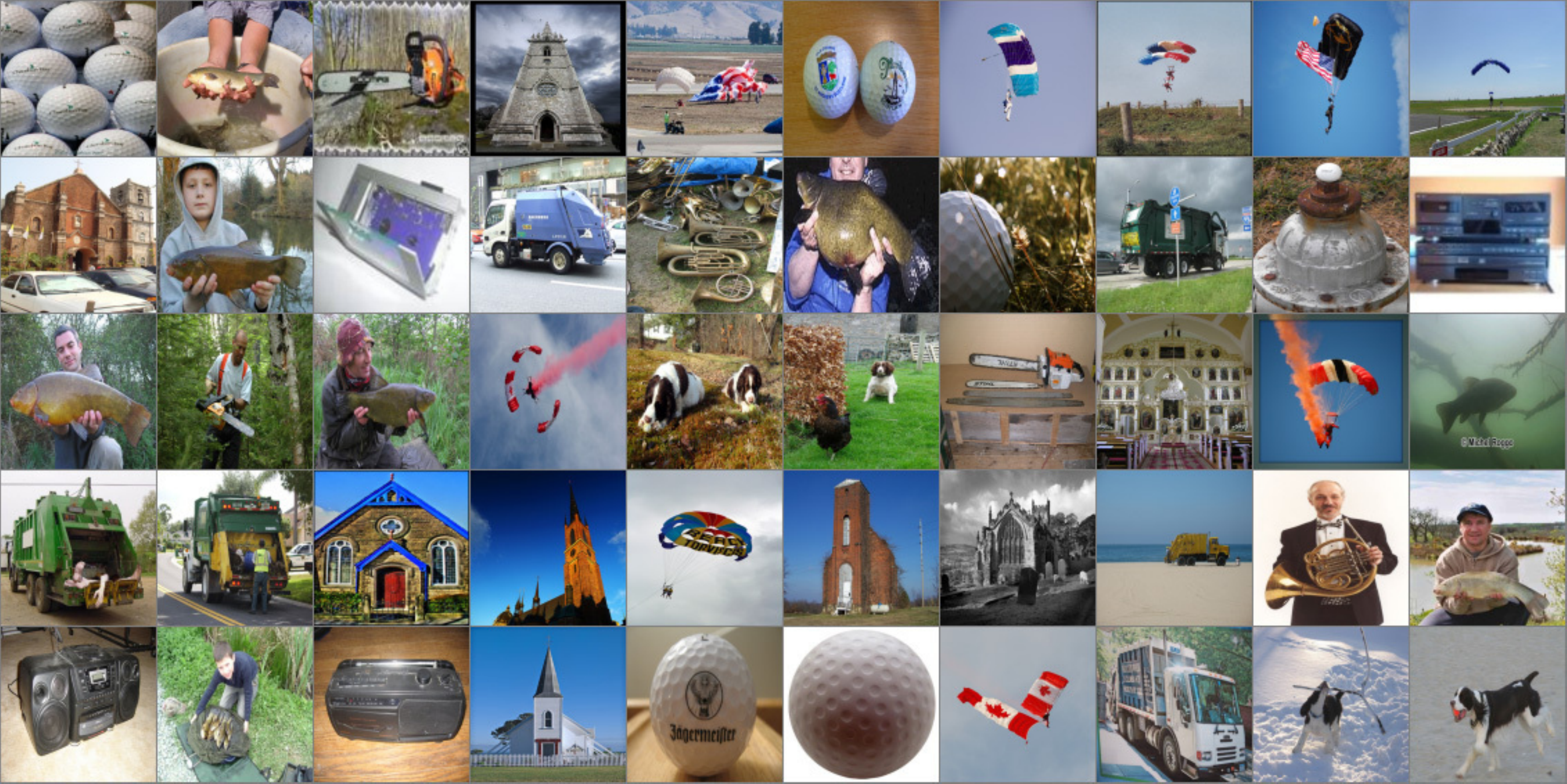}};
            \node[anchor=east] (anchors) at (pic.west){\includegraphics[height=3.8cm]{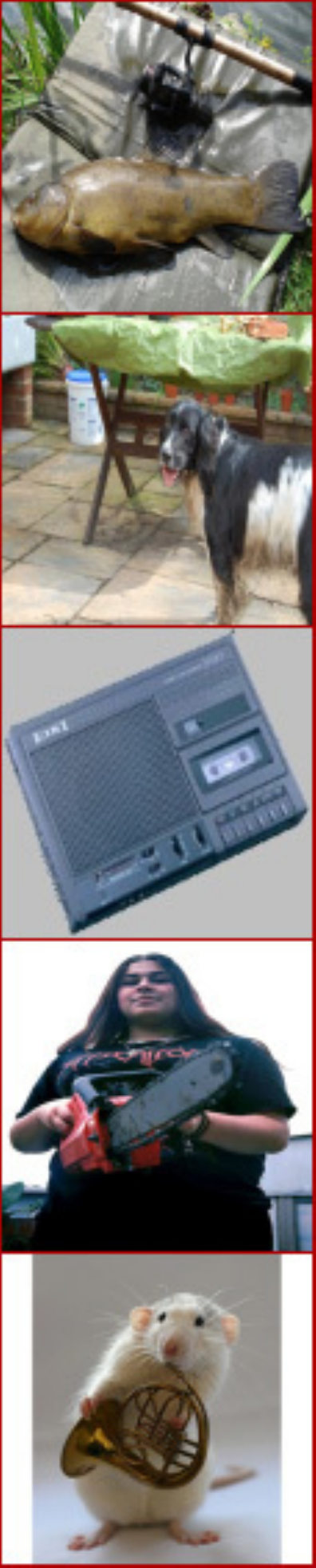}};
        \end{tikzpicture}
        \subcaption{Imagenette, $z_\text{style}$}\label{fig:nn_vae_imagenette}
    \end{minipage}%
    \begin{minipage}{0.5\linewidth}
        \centering
        \begin{tikzpicture}
            \node (pic) at (0, 0){\includegraphics[height=3.8cm]{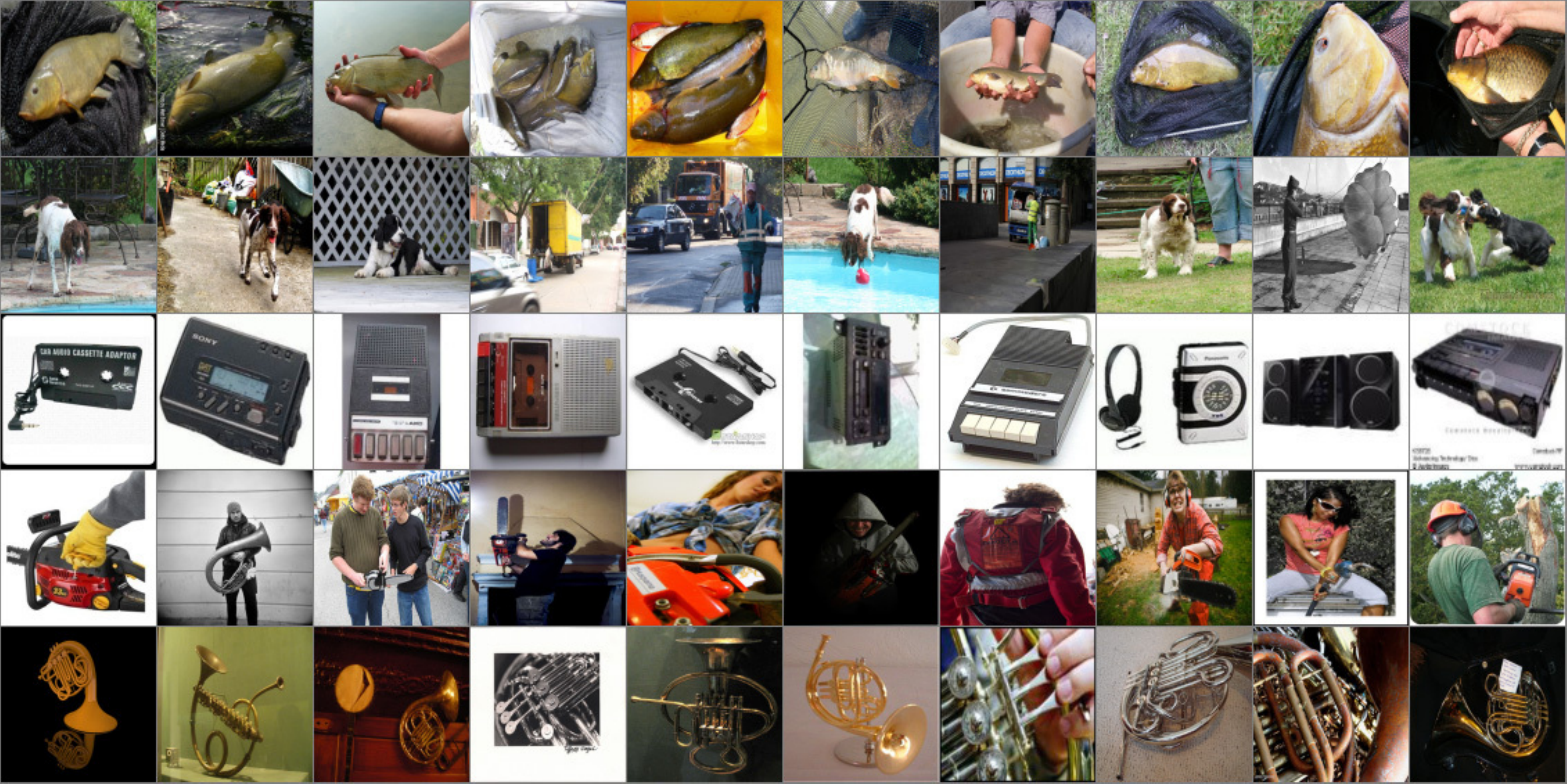}};
            \node[anchor=east] (anchors) at (pic.west){\includegraphics[height=3.8cm]{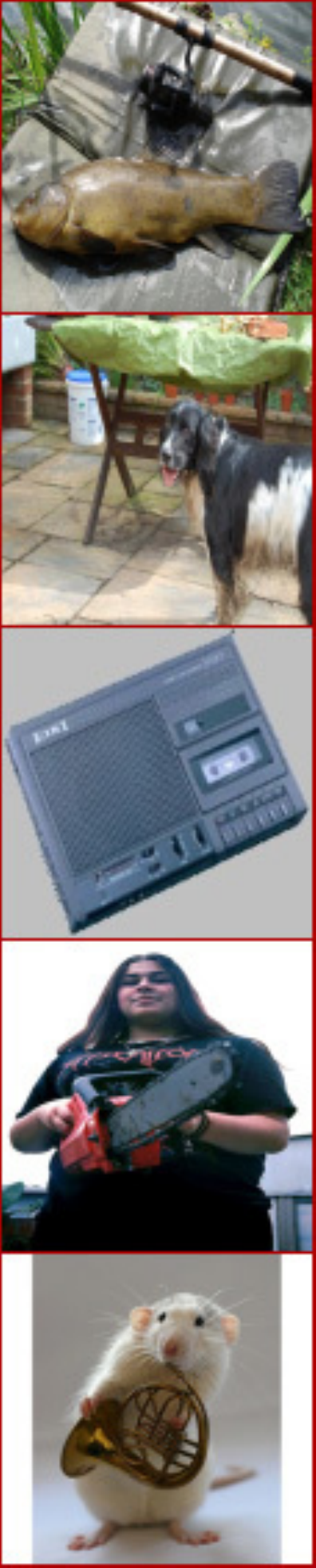}};
        \end{tikzpicture}
        \subcaption{Imagenette, $z_\text{content}$}\label{fig:nn_cl_imagenette}
    \end{minipage}
    \vspace{0.2cm}

    \begin{minipage}{0.5\linewidth}
        \centering
        \begin{tikzpicture}
            \node (pic) at (0, 0){\includegraphics[height=3.8cm]{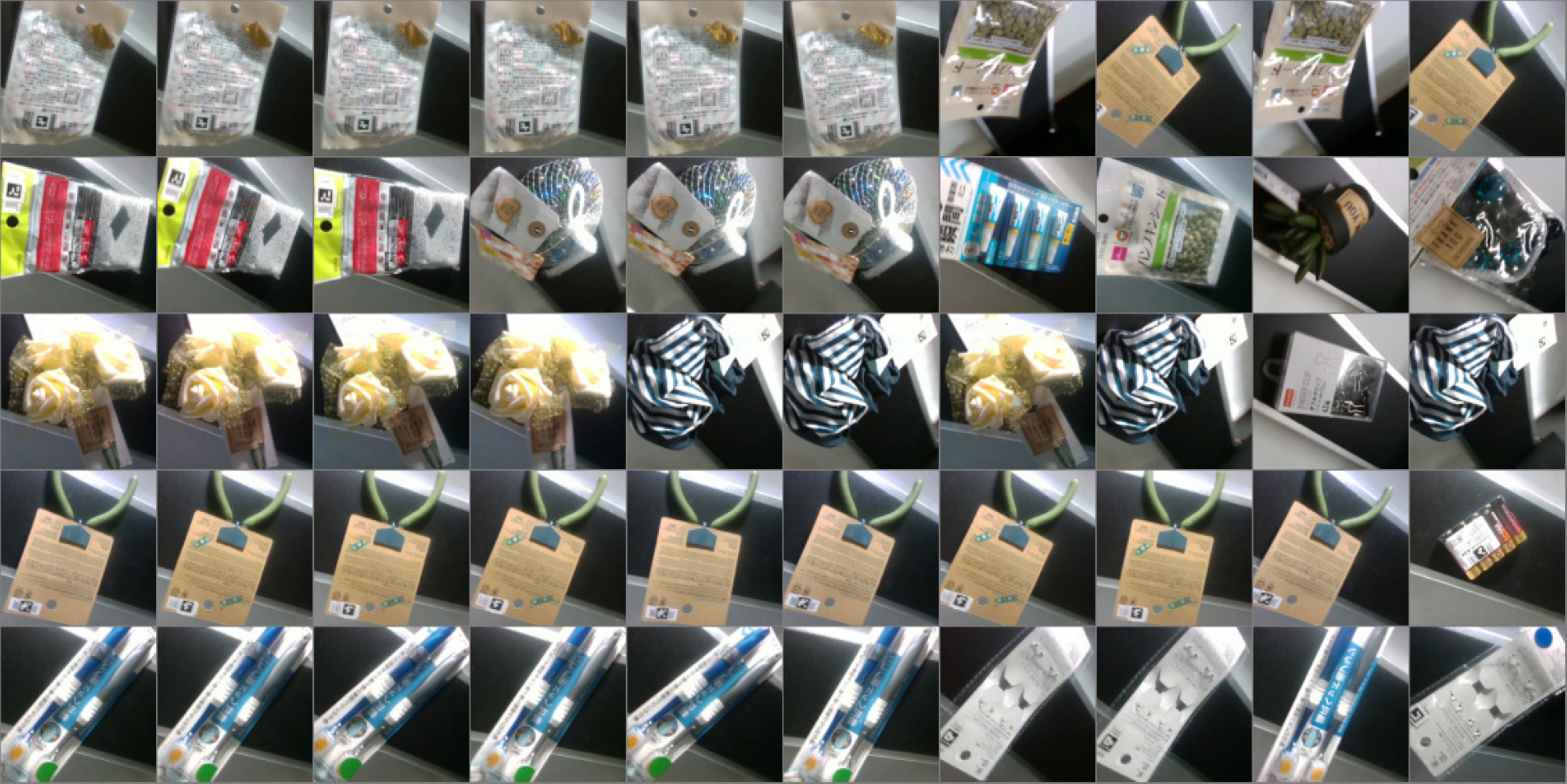}};
            \node[anchor=east] (anchors) at (pic.west){\includegraphics[height=3.8cm]{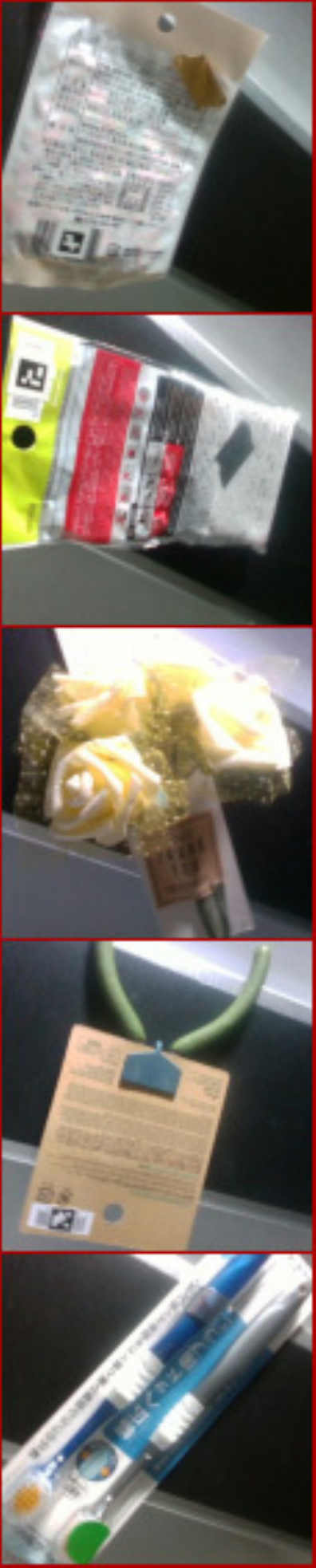}};
        \end{tikzpicture}
        \subcaption{DAISO-100, $z_\text{style}$}\label{fig:nn_vae_daiso100}
    \end{minipage}%
    \begin{minipage}{0.5\linewidth}
        \centering
        \begin{tikzpicture}
            \node (pic) at (0, 0){\includegraphics[height=3.8cm]{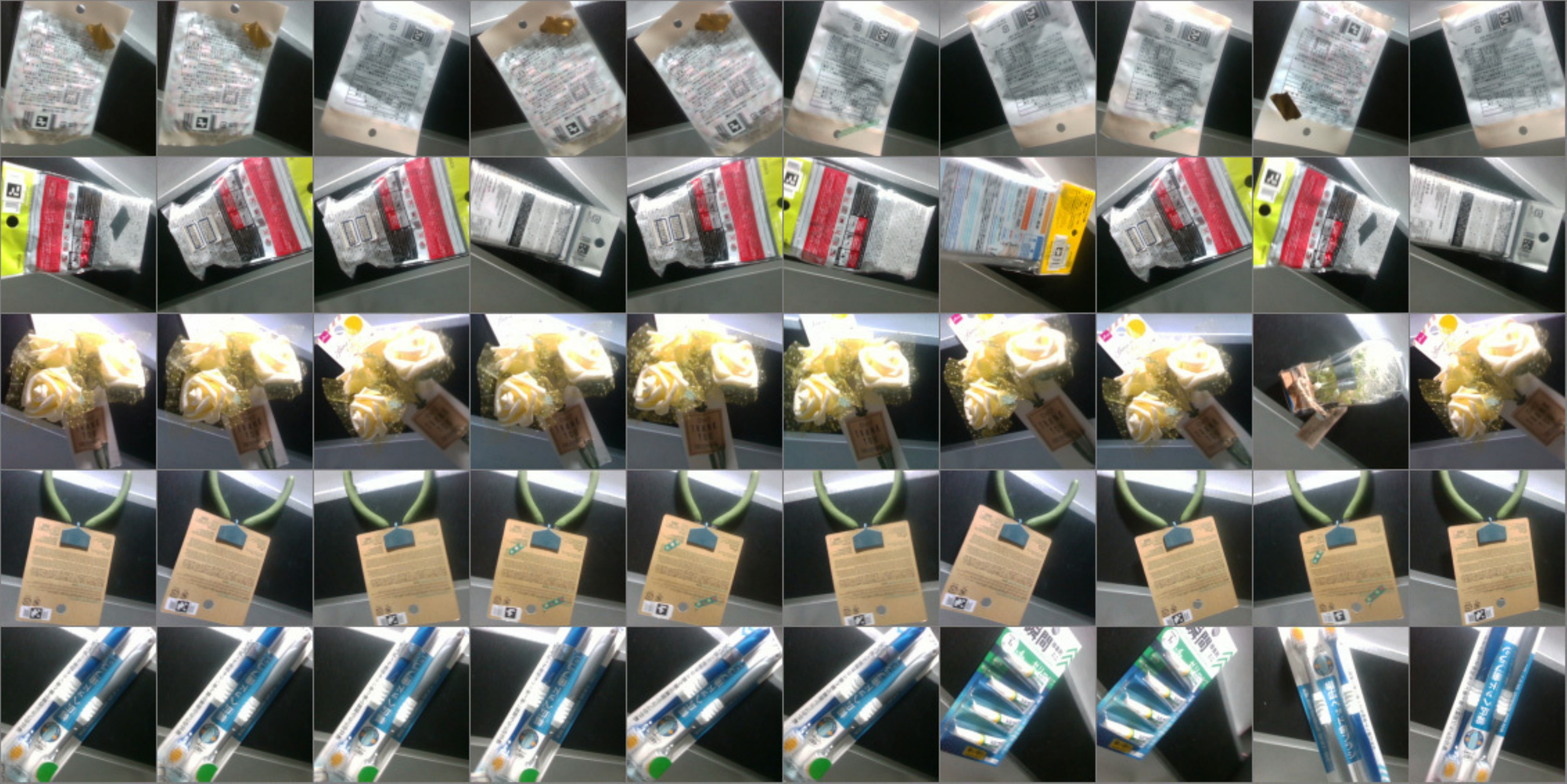}};
            \node[anchor=east] (anchors) at (pic.west){\includegraphics[height=3.8cm]{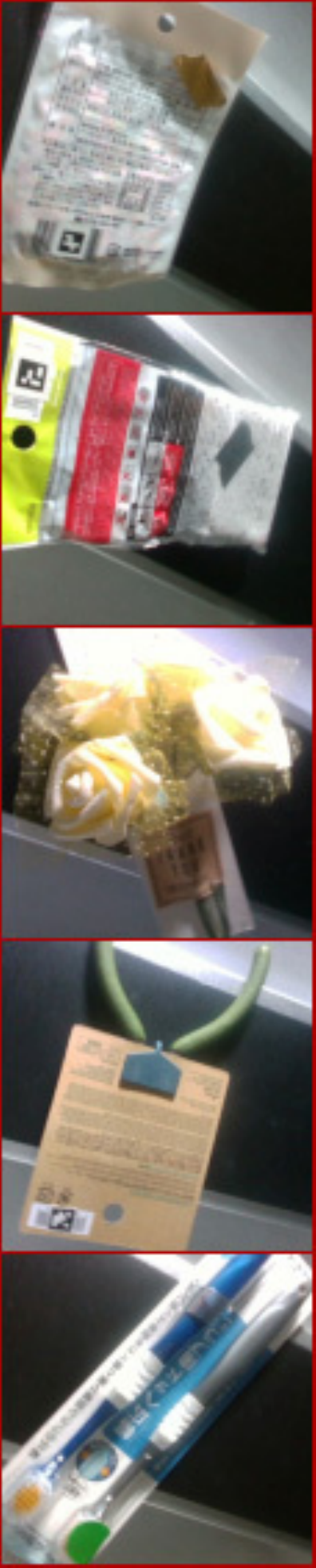}};
        \end{tikzpicture}
        \subcaption{DAISO-100, $z_\text{content}$}\label{fig:nn_cl_daiso100}
    \end{minipage}
    \caption{%
    Visualization of the neighbors in terms of $z_\text{style}$ and $z_\text{content}$.
    The leftmost red framed data are anchors, and their neighbors are listed in order of distance.
    }
    \label{fig:nn}
\end{figure*}

We illustrate the neighbors of some example test data in \cref{fig:nn}.
The neighbors had similar styles in the style feature spaces shown in \cref{fig:nn_vae_mnist,fig:nn_vae_font}.
However, for MNIST, the neighbors tended to simply be similar data.
For the Google Fonts dataset, the neighbors had different characters with similar bounding boxes.
Changes in bounding boxes were observed as the styles in the style-transfer experiments.
In style-independent feature spaces, the neighbors had almost the same content in different styles, as shown in \cref{fig:nn_cl_mnist,fig:nn_cl_font}.
This result agrees with the characteristics of the CL; the CL models extract data augmentation-independent features and are suitable for classification~\cite{SimCLR,MoCo,SimSiam}.

\cref{fig:nn_vae_imagenette,fig:nn_cl_imagenette} show the results on Imagenette.
\cref{fig:nn_vae_imagenette} is difficult to interpret; however, \cref{fig:nn_cl_imagenette} illustrates that CL mapped images with the same object to similar vectors.
\cref{fig:nn_vae_daiso100,fig:nn_cl_daiso100} show the results on DAISO-100.
For CL model, the results were similar to those of Imagenette; the neighbors were almost the same object in various styles.
\cref{fig:nn_vae_daiso100} shows that the CVAE extracted similar features corresponding to the styles.
Specifically, we can see that the neighbors were with different objects and similar backgrounds.

\subsection{Combination with various CL Methods}
Although we evaluated the proposed method combined with MoCo v2~\cite{chen2020mocov2}, we also examined the combinations with other CL methods.
We used SimCLR~\cite{SimCLR}, SimSiam~\cite{SimSiam}, and VICReg~\cite{VICReg}.
SimCLR is a classical CL method on which MoCo~\cite{MoCo,chen2020mocov2} is based.
SimSiam and VICReg are noncontrastive CL methods, and they do not use negative samples while learning.
As a reference, we tested a supervised variant of the proposed method.
The one-hot encoded class labels are used instead of the CL feature vectors.
We compared the effects of the different CL methods using the same experiments described in \cref{sec:da_removal_mnist,sec:interp_mnist}.

The hyperparameters of the CL methods were set to their defaults described in the original papers.
For training CVAE, we used the same hyperparameters as that in the MoCo v2 version except for the combination of SimSiam and MNIST.
In the experiment with this combination, we changed the weight of the MI $\lambda_\text{MINE}$ to $0.1$ because the styles were not extracted well with the original setting.

\begin{figure*}
    \centering
    \begin{minipage}{\linewidth}
        \centering
        \begin{tikzpicture}
            \node (pic) at (0, 0) {\includegraphics[height=3cm]{fig/da_removal_mnist.pdf}};
            \node[anchor=north east, yshift=-0.3cm] at (pic.north west) {Input $x$};
            \node[anchor=north east, yshift=-1cm] at (pic.north west) {Data augmented input};
            \node[anchor=north east, yshift=-1.7cm] at (pic.north west) {Reconstruction $D_\theta(z_\text{style}, z_\text{content})$};
            \node[anchor=north east, yshift=-2.5cm] at (pic.north west) {Style-free reconstruction $D_\theta(\mathbf{0}, z_\text{content})$};
        \end{tikzpicture}
        \subcaption{MNIST, MoCo v2 (from \cref{fig:da_removal_mnist})}
    \end{minipage}
    \vspace{0.2cm}

    \begin{minipage}{0.5\linewidth}
        \centering
        \includegraphics[width=0.9\linewidth]{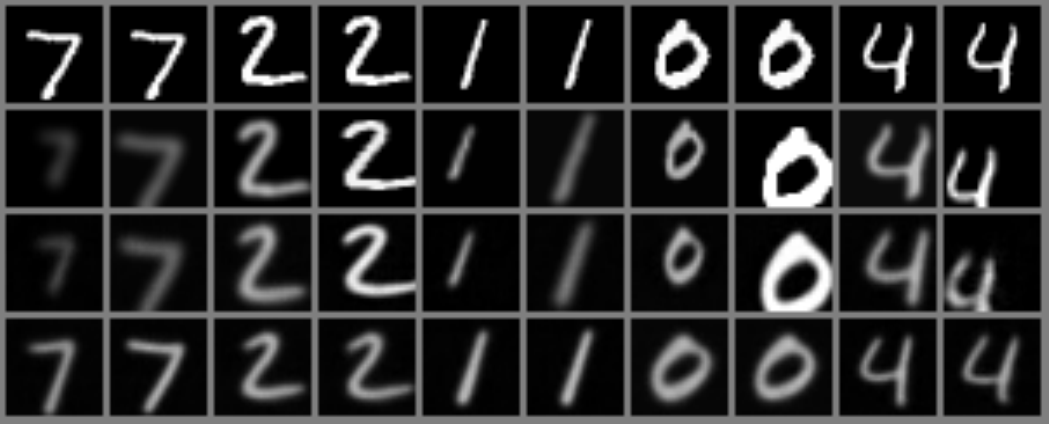}
        \subcaption{MNIST, SimCLR~\cite{SimCLR}}\label{fig:da_removal_mnist_simclr}
    \end{minipage}%
    \begin{minipage}{0.5\linewidth}
        \centering
        \includegraphics[width=0.9\linewidth]{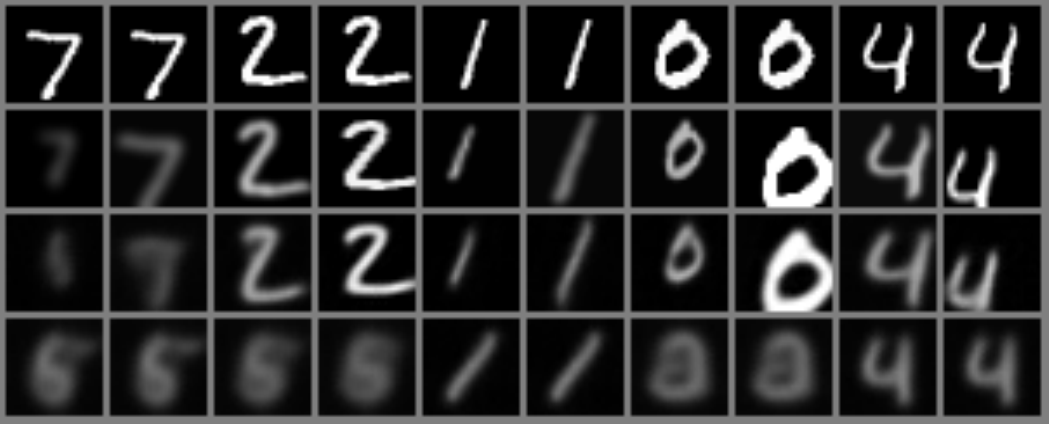}
        \subcaption{MNIST, SimSiam~\cite{SimSiam}}\label{fig:da_removal_mnist_simsiam}
    \end{minipage}%
    \vspace{0.2cm}

    \begin{minipage}{0.5\linewidth}
        \centering
        \includegraphics[width=0.9\linewidth]{fig/da_removal_mnist_simclr.pdf}
        \subcaption{MNIST, VICReg~\cite{VICReg}}\label{fig:da_removal_mnist_vicreg}
    \end{minipage}%
    \begin{minipage}{0.5\linewidth}
        \centering
        \includegraphics[width=0.9\linewidth]{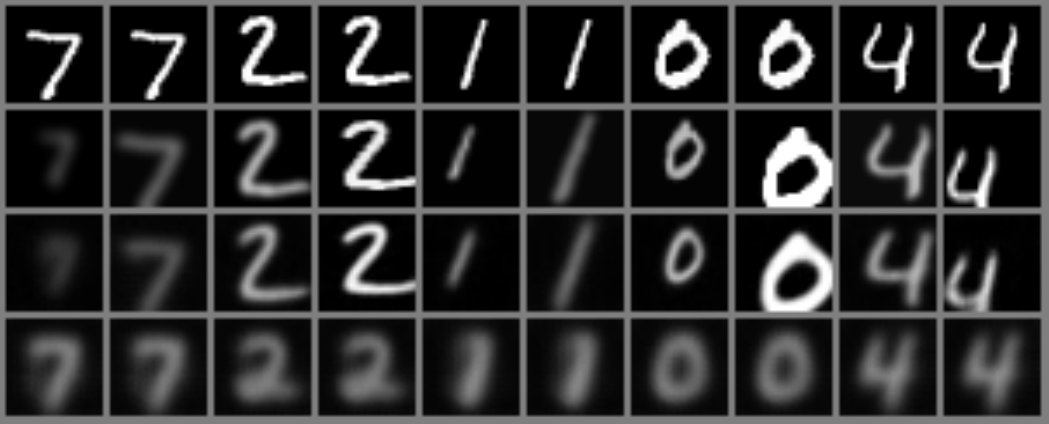}
        \subcaption{MNIST, supervised (reference)}\label{fig:da_removal_mnist_supervised}
    \end{minipage}%
    \vspace{0.2cm}

    \begin{minipage}{\linewidth}
        \centering
        \begin{tikzpicture}
            \node (pic) at (0, 0) {\includegraphics[height=3cm]{fig/da_removal_font.pdf}};
            \node[anchor=north east, yshift=-0.3cm] at (pic.north west) {Input $x$};
            \node[anchor=north east, yshift=-1cm] at (pic.north west) {Data augmented input};
            \node[anchor=north east, yshift=-1.7cm] at (pic.north west) {Reconstruction $D_\theta(z_\text{style}, z_\text{content})$};
            \node[anchor=north east, yshift=-2.5cm] at (pic.north west) {Style-free reconstruction $D_\theta(\mathbf{0}, z_\text{content})$};
        \end{tikzpicture}
        \subcaption{Google Fonts, MoCo v2 (from \cref{fig:da_removal_font})}
    \end{minipage}
    \vspace{0.2cm}

    \begin{minipage}{0.5\linewidth}
        \centering
        \includegraphics[width=0.9\linewidth]{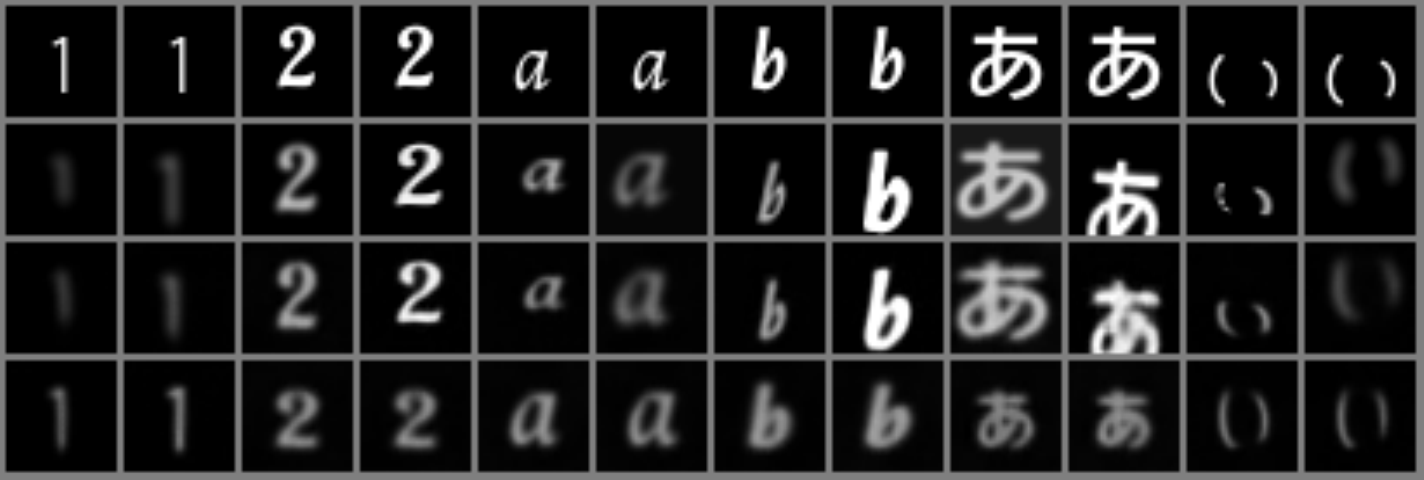}
        \subcaption{Google Fonts, SimCLR~\cite{SimCLR}}\label{fig:da_removal_font_simclr}
    \end{minipage}%
    \begin{minipage}{0.5\linewidth}
        \centering
        \includegraphics[width=0.9\linewidth]{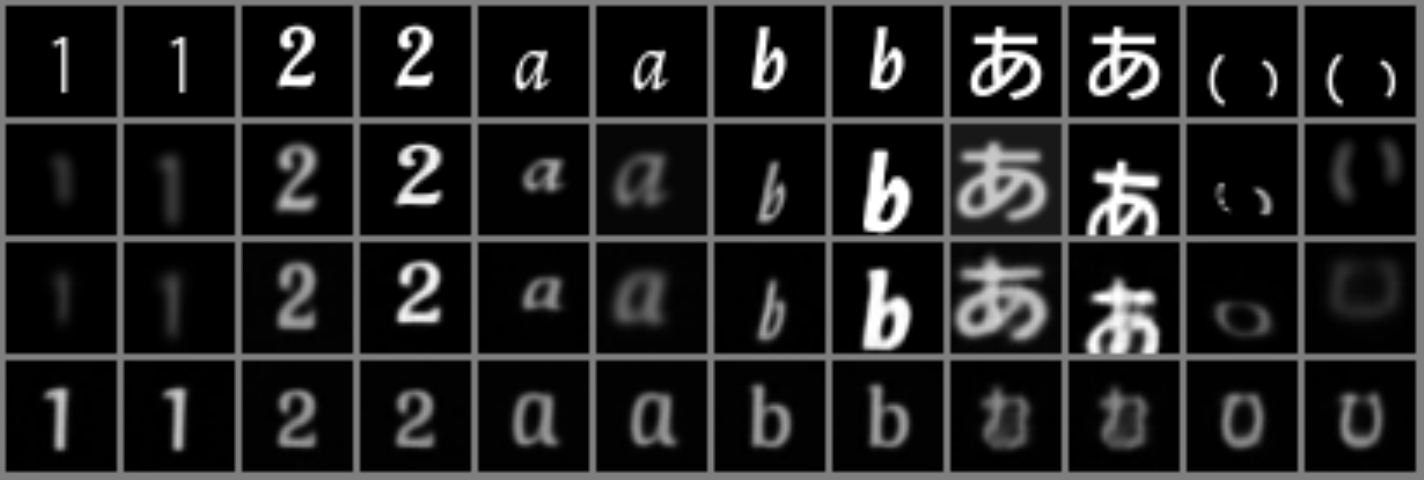}
        \subcaption{Google Fonts, SimSiam~\cite{SimSiam}}\label{fig:da_removal_font_simsiam}
    \end{minipage}%
    \vspace{0.3cm}

    \begin{minipage}{0.5\linewidth}
        \centering
        \includegraphics[width=0.9\linewidth]{fig/da_removal_font_simclr.pdf}
        \subcaption{Google Fonts, VICReg~\cite{VICReg}}\label{fig:da_removal_font_vicreg}
    \end{minipage}%
    \begin{minipage}{0.5\linewidth}
        \centering
        \includegraphics[width=0.9\linewidth]{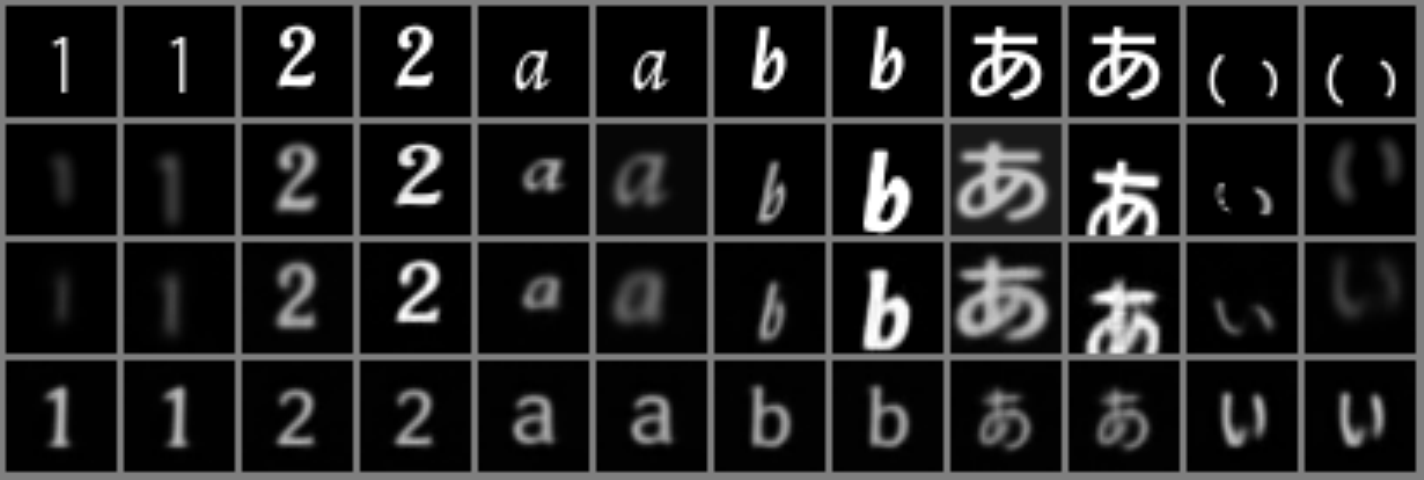}
        \subcaption{Google Fonts, supervised (reference)}\label{fig:da_removal_font_supervised}
    \end{minipage}%
    \caption{%
    Examples of isolating and capturing data augmentation features when the proposed method is combined with different CL methods.
    From top to bottom: original images, images after data augmentation, reconstructed images $D_\theta(z_\text{style}, z_\text{content})$, and images reconstructed without style features $D_\theta(\mathbf{0}, z_\text{content})$.
    }
    \label{fig:da_removal_variations}
\end{figure*}

\cref{fig:da_removal_variations} shows the results of data augmentation isolation.
Data augmentation features were isolated by the CL methods, and the CVAE captured them, regardless of the selected CL method.
The reconstruction images without styles were blurry in the experiment using SimSiam with MNIST (bottom row of \cref{fig:da_removal_mnist_simsiam}).
This indicates that style extraction performance depends on how the CL method is trained.
Besides, when class labels are given instead of the CL feature, the reconstruction images without styles looked similar to the average of the class (bottom row of \cref{fig:da_removal_mnist_supervised,fig:da_removal_font_supervised}).
These results show that the proposed method successfully captured features that the condition of the CVAE (i.e., style-independent CL feature $z_\text{content}$ or the class label) do not contain.

\begin{figure*}
    \centering
    \begin{minipage}{0.5\linewidth}
        \centering
        \begin{tikzpicture}
            \node (pic) at (0, 0) {\includegraphics[width=0.8\linewidth]{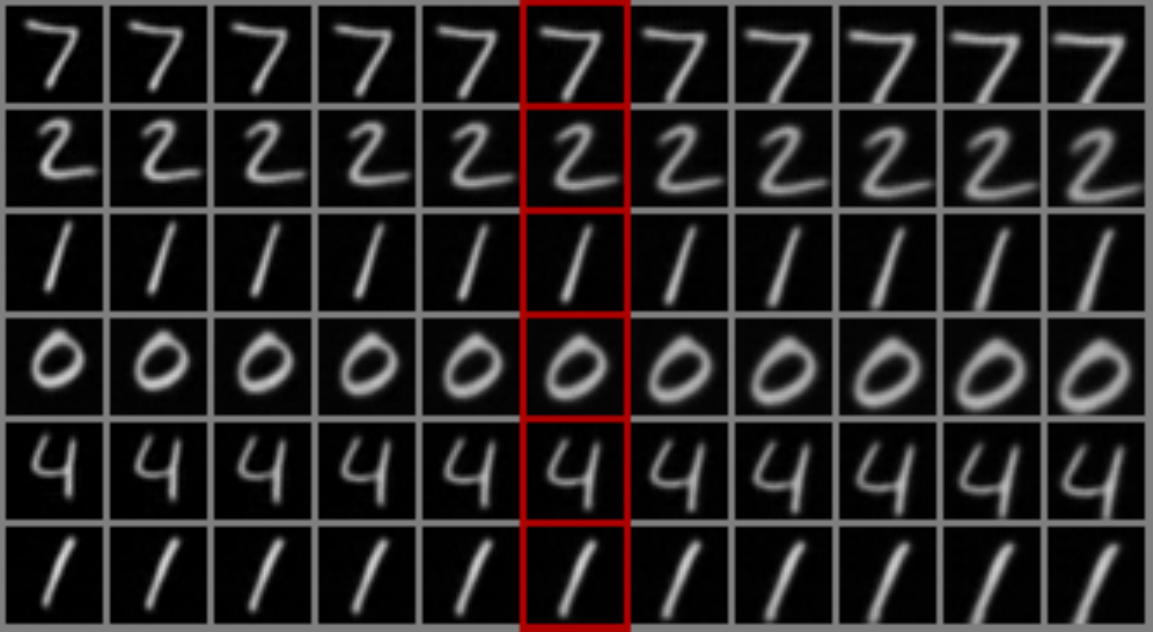}};
            \node[anchor=south east, xshift=1cm] at (pic.north west) {$z_\text{content}$};
            \node[anchor=north west, xshift=-0.8cm] at (pic.south east) {$z_\text{style}$};
            \draw[-latex](pic.south west) -- (pic.north west);
            \draw[-latex](pic.south west) -- (pic.south east);
        \end{tikzpicture}
        \subcaption{MNIST, SimCLR~\cite{SimCLR}}\label{fig:interp_mnist_simclr}
    \end{minipage}%
    \begin{minipage}{0.5\linewidth}
        \centering
        \begin{tikzpicture}
            \node (pic) at (0, 0) {\includegraphics[width=0.8\linewidth]{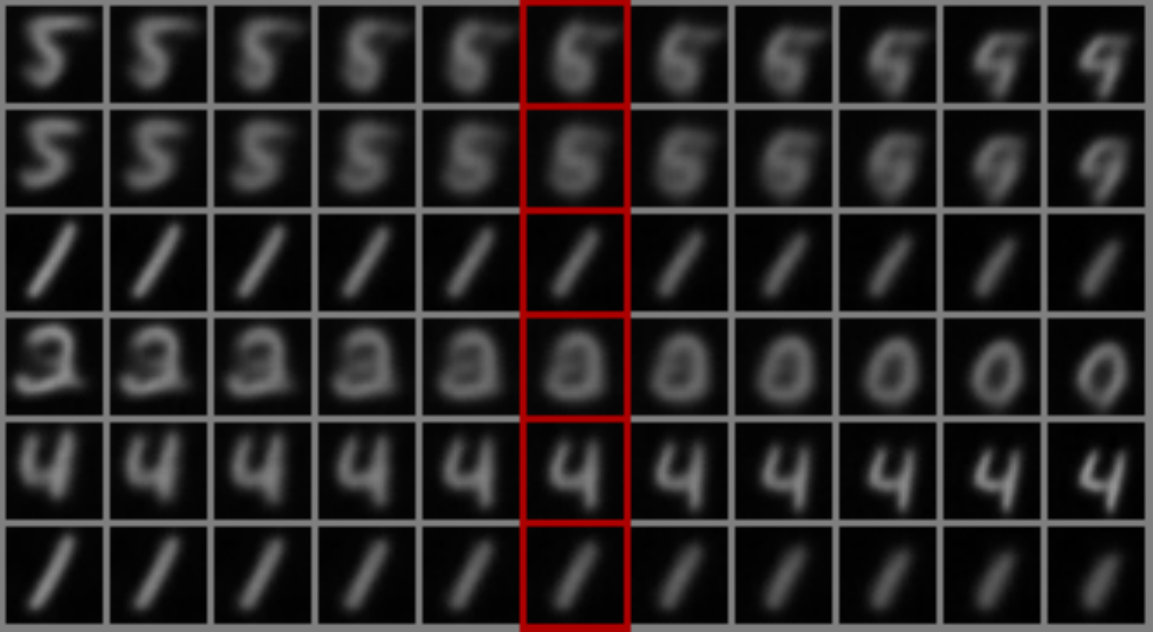}};
            \node[anchor=south east, xshift=1cm] at (pic.north west) {$z_\text{content}$};
            \node[anchor=north west, xshift=-0.8cm] at (pic.south east) {$z_\text{style}$};
            \draw[-latex](pic.south west) -- (pic.north west);
            \draw[-latex](pic.south west) -- (pic.south east);
        \end{tikzpicture}
        \subcaption{MNIST, SimSiam~\cite{SimSiam}}\label{fig:interp_mnist_simsiam}
    \end{minipage}%
    \vspace{0.2cm}

    \begin{minipage}{0.5\linewidth}
        \centering
        \begin{tikzpicture}
            \node (pic) at (0, 0) {\includegraphics[width=0.8\linewidth]{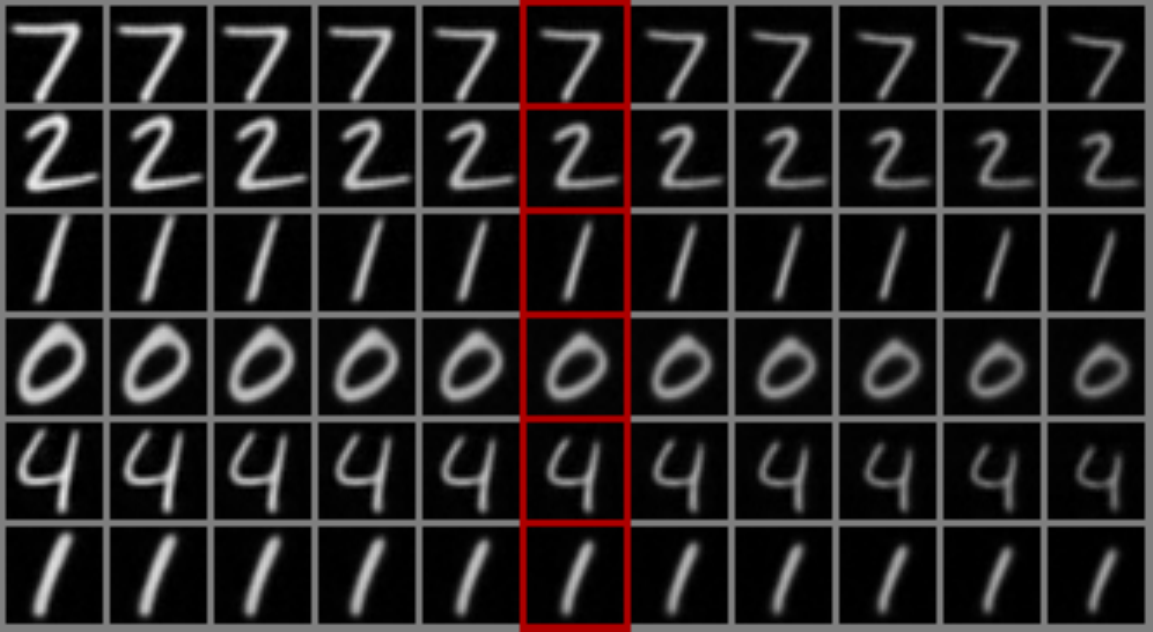}};
            \draw[-latex](pic.south west) -- (pic.north west);
            \draw[-latex](pic.south west) -- (pic.south east);
        \end{tikzpicture}
        \subcaption{MNIST, VICReg~\cite{VICReg}}\label{fig:interp_mnist_vicreg}
    \end{minipage}%
    \begin{minipage}{0.5\linewidth}
        \centering
        \begin{tikzpicture}
            \node (pic) at (0, 0) {\includegraphics[width=0.8\linewidth]{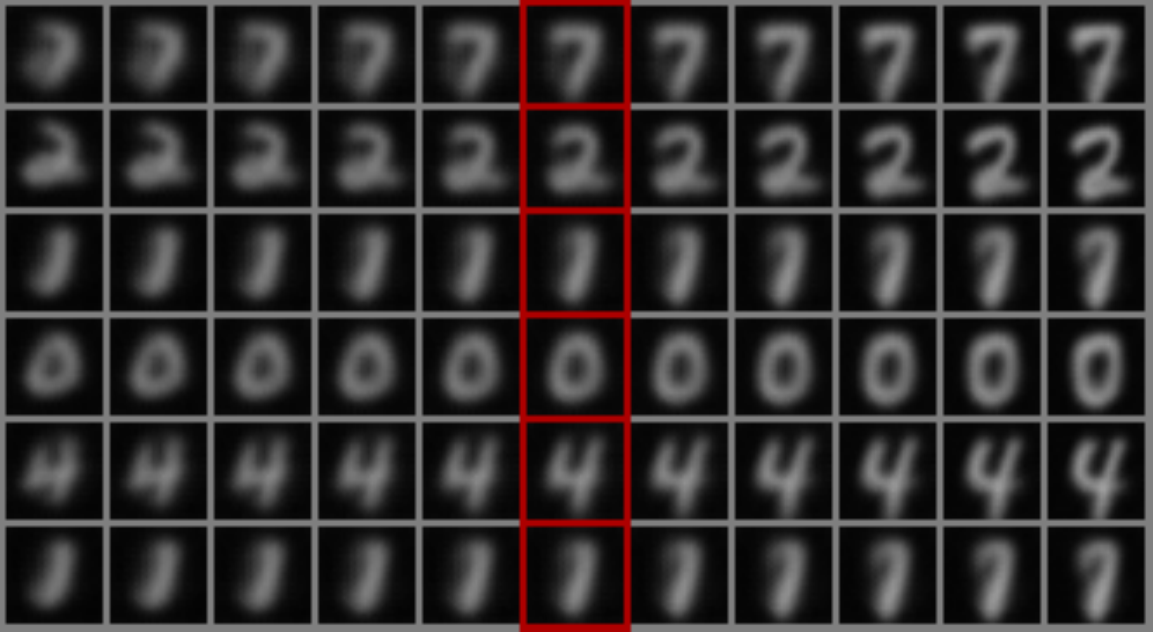}};
            \node[anchor=south east, xshift=1cm] at (pic.north west) {$z_\text{content}$};
            \node[anchor=north west, xshift=-0.8cm] at (pic.south east) {$z_\text{style}$};
            \draw[-latex](pic.south west) -- (pic.north west);
            \draw[-latex](pic.south west) -- (pic.south east);
        \end{tikzpicture}
        \subcaption{MNIST, supervised (reference)}\label{fig:interp_mnist_supervised}
    \end{minipage}%
    \vspace{0.2cm}

    \begin{minipage}{0.5\linewidth}
        \centering
        \begin{tikzpicture}
            \node (pic) at (0, 0) {\includegraphics[width=0.8\linewidth]{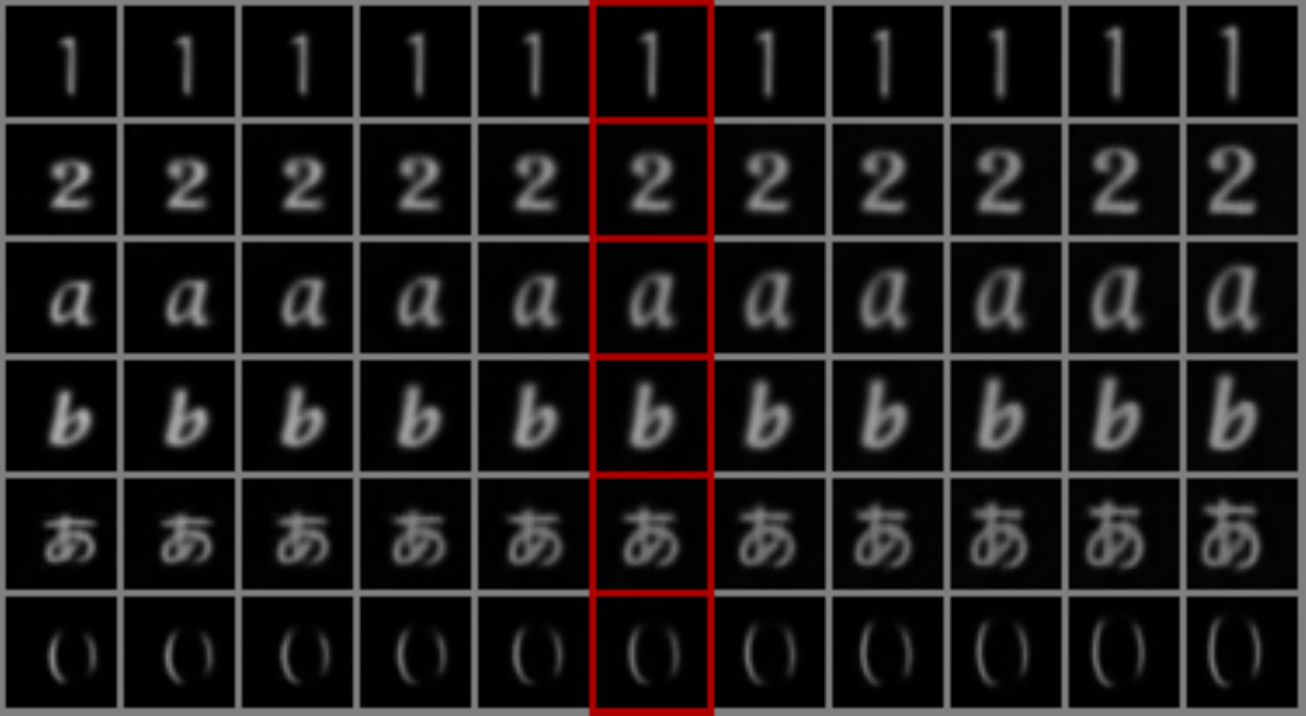}};
            \node[anchor=south east, xshift=1cm] at (pic.north west) {$z_\text{content}$};
            \node[anchor=north west, xshift=-0.8cm] at (pic.south east) {$z_\text{style}$};
            \draw[-latex](pic.south west) -- (pic.north west);
            \draw[-latex](pic.south west) -- (pic.south east);
        \end{tikzpicture}
        \subcaption{Google Fonts, SimCLR~\cite{SimCLR}}\label{fig:interp_font_simclr}
    \end{minipage}%
    \begin{minipage}{0.5\linewidth}
        \centering
        \begin{tikzpicture}
            \node (pic) at (0, 0) {\includegraphics[width=0.8\linewidth]{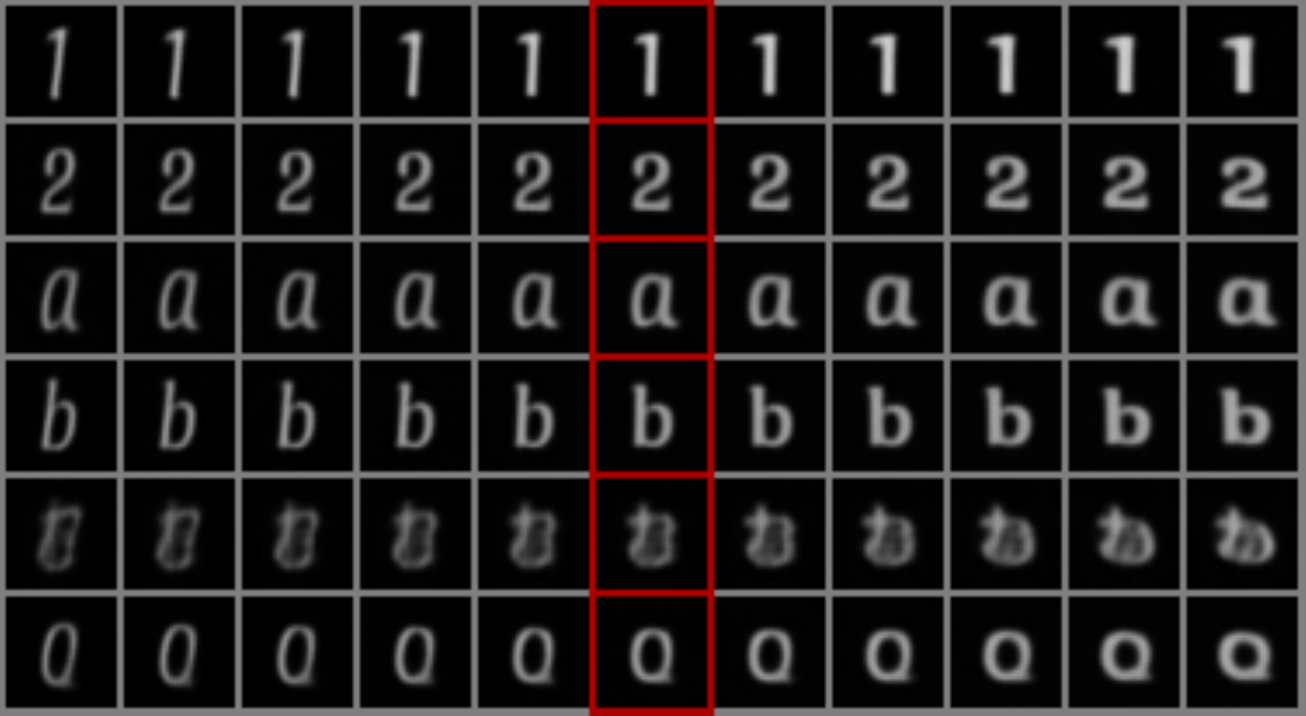}};
            \node[anchor=south east, xshift=1cm] at (pic.north west) {$z_\text{content}$};
            \node[anchor=north west, xshift=-0.8cm] at (pic.south east) {$z_\text{style}$};
            \draw[-latex](pic.south west) -- (pic.north west);
            \draw[-latex](pic.south west) -- (pic.south east);
        \end{tikzpicture}
        \subcaption{Google Fonts, SimSiam~\cite{SimSiam}}\label{fig:interp_font_simsiam}
    \end{minipage}%
    \vspace{0.2cm}

    \begin{minipage}{0.5\linewidth}
        \centering
        \begin{tikzpicture}
            \node (pic) at (0, 0) {\includegraphics[width=0.8\linewidth]{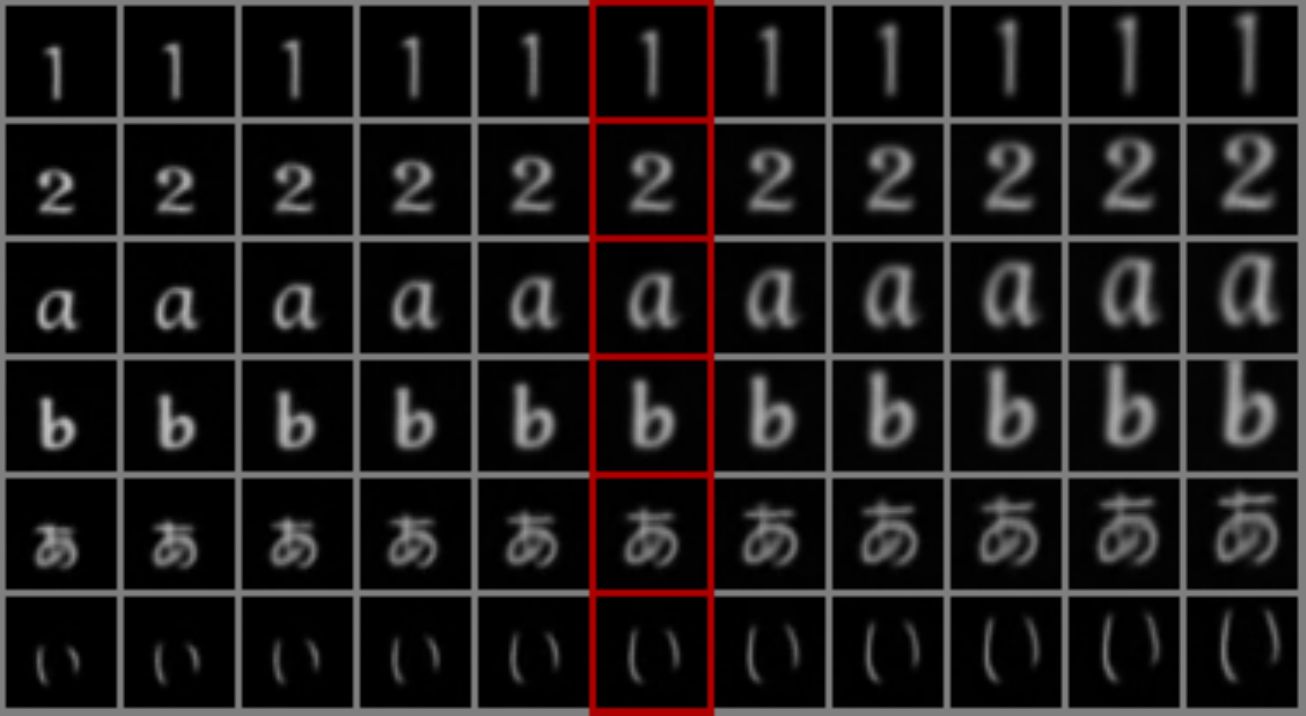}};
            \node[anchor=south east, xshift=1cm] at (pic.north west) {$z_\text{content}$};
            \node[anchor=north west, xshift=-0.8cm] at (pic.south east) {$z_\text{style}$};
            \draw[-latex](pic.south west) -- (pic.north west);
            \draw[-latex](pic.south west) -- (pic.south east);
        \end{tikzpicture}
        \subcaption{Google Fonts, VICReg~\cite{VICReg}}\label{fig:interp_font_vicreg}
    \end{minipage}%
    \begin{minipage}{0.5\linewidth}
        \centering
        \begin{tikzpicture}
            \node (pic) at (0, 0) {\includegraphics[width=0.8\linewidth]{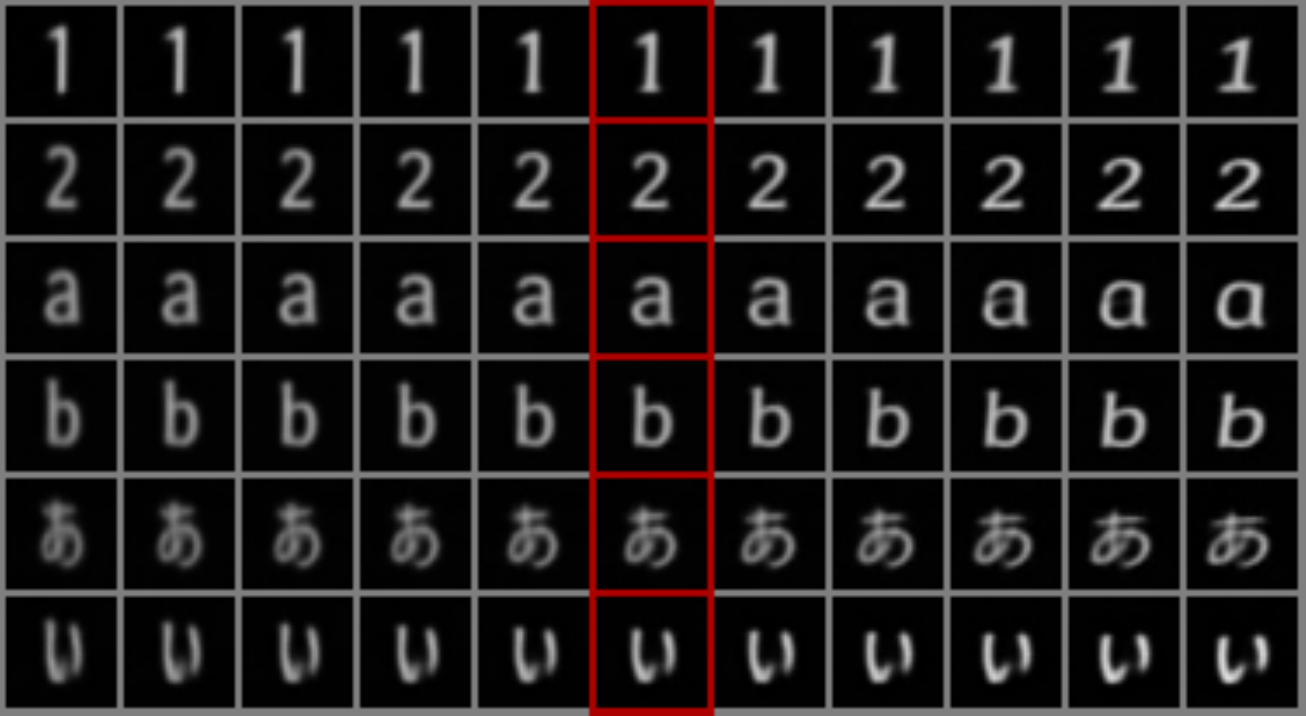}};
            \node[anchor=south east, xshift=1cm] at (pic.north west) {$z_\text{content}$};
            \node[anchor=north west, xshift=-0.8cm] at (pic.south east) {$z_\text{style}$};
            \draw[-latex](pic.south west) -- (pic.north west);
            \draw[-latex](pic.south west) -- (pic.south east);
        \end{tikzpicture}
        \subcaption{Google Fonts, supervised (reference)}\label{fig:interp_font_supervised}
    \end{minipage}%
    \caption{%
    Examples of conditional generation when the proposed method is combined with different CL methods.
    Each row corresponds to different $z_\text{content}$ from some test data.
    Each columns corresponds to interpolated $z_\text{style} (0 \leq \|z_\text{style}\| \leq 3)$ along a randomly chosen unit vector.
    The images of the center column with the red frame are generated without the style features (i.e., $z_\text{style}=\mathbf{0}$).
    }
    \label{fig:interp_variations}
\end{figure*}

The results of conditional generation are shown in \cref{fig:interp_variations}.
Note that the learned feature spaces of the CVAE are different in every experiment, and therefore, the changes in styles were not the same across the experiments.
\cref{fig:interp_variations} illustrates that the CVAE can extract style features with any CL method successfully.
The supervised variant on the Google Font dataset (\cref{fig:interp_mnist_supervised,fig:interp_font_supervised}) generated more variation in styles by $z_\text{style}$.
Specifically, the changes in the font faces were captured as shown in \cref{fig:interp_font_supervised}.

%% file: discussion.tex
\section{Discussion}
In our experiments using MNIST-like datasets, the proposed method successfully extracted style features (\cref{fig:da_removal,fig:interp,fig:style_transfer,fig:nn}).
The extracted styles were related to the characters' size, thickness, slant, position, etc.
Additionally, the style-extraction performance was retained across different CL methods (\cref{fig:da_removal_variations,fig:interp_variations}).

However, in the experiments with the Google Fonts dataset, font faces that can be easily understood as styles are not extracted as styles (\cref{fig:da_removal_font,fig:interp_font,fig:style_transfer_font,fig:nn_vae_font}).
These results can be attributed to the data augmentation used.
The proposed method aims to capture information that the CL isolates, and the information isolated by the CL heavily relies on data augmentation.
The data augmentation was appropriate for isolating the styles in the case of MNIST, but not for the Google Fonts dataset.
For capturing font faces as styles, data augmentation should cover differences between different font faces (e.g.\ by adding morphological transformations).

The choice of the condition for the CVAE is also important.
As indicated by the experiments with the different CL methods and the supervised variant (\cref{fig:da_removal_variations,fig:interp_variations}), the extracted style features are slightly different across the CL methods.
Moreover, the proposed method extracted more diverse styles when the class labels were provided (i.e.,\ the supervised variant).
These results suggest that style extraction performance of the proposed method depends on the quality of the style-independent feature $z_\text{content}$.
We can change the conditions for the CVAE depending on the styles and datasets we want to extract.
We used CL as the condition to train the model with completely unlabeled data.
However, if some labels are available, making use of them in a supervised or semi-supervised manner is a possible choice.

The proposed method could also deal with real-world natural image datasets in the experiments (\cref{fig:style_transfer_imagenette,fig:style_transfer_daiso100,fig:nn_vae_imagenette,fig:nn_cl_imagenette,fig:nn_vae_daiso100,fig:nn_cl_daiso100}).
The style-transfer experiments indicated that the CVAE extracted styles.
However, in the Imagenette experiments, we did not find common styles in the neighbor analysis, although we did find them in the DAISO-100 experiments.
This can be attributed to Imagenette being a small dataset with a relatively large variety of data, which resulted in the learned style feature space being too sparse to find meaningful neighbors.

In terms of the natural image datasets, the quality of the generated images is a limitation.
Our aim is not to generate images but to extract style features; however, high-quality image generation is desirable for evaluating the learned features.
Improving the decoder model or combining it with adversarial models such as GANs~\cite{GAN} is one approach to achieve this.

Although the experimental results illustrated the style-extracting ability of the proposed method, the evaluations were qualitative.
We need quantitative evaluations for comparing the performance of the different CL methods in detail.
One approach to achieve this is by preparing a dataset with explicit style labels and a set of data augmentation operations that can perturb only the explicit styles.

%% file: conclusions.tex
\section{Conclusions}
In this paper, we proposed a style feature extracting CVAE conditioned by CL.
The CL models isolate the style features, and the proposed CVAE captures them.
Additionally, we introduced a constraint based on mutual information to aid the CVAE in extracting features that correspond exclusively to styles.
Our experiments used two simple datasets, i.e.,\ MNIST and the Google Fonts dataset, and showed that the proposed method effectively extracts the style features.
Additional experiments on larger, real-world image datasets, i.e.,\ Imagenette and DAISO-100, further demonstrated that the proposed method could effectively deal with larger-scale datasets.

Although the proposed method was successful in extracting style information, there is room for improvement.
The decoder can be improved to better assess the learned style features when using natural image datasets.
One potential approach to achieve this would be to combine the method with GANs~\cite{GAN}.
Another area for improvement would be to extend the method to other domains such as time-series data, for example, audio, speech, and biosignals, which have styles corresponding to subjects and for which CL has been reported to be effective~\cite{Saeed2021,Eldele2021}.